\documentclass[10pt]{article} % For LaTeX2e
\usepackage[preprint]{tmlr}
\usepackage[flushleft]{threeparttable}
\usepackage{amsmath,amsfonts,bm}

\def\eqref#1{equation~\ref{#1}}
\def\1{\bm{1}}

\DeclareMathAlphabet{\mathsfit}{\encodingdefault}{\sfdefault}{m}{sl}
\SetMathAlphabet{\mathsfit}{bold}{\encodingdefault}{\sfdefault}{bx}{n}

\usepackage{hyperref}
\usepackage{url}
\usepackage{booktabs,graphicx,amsmath,amssymb,multirow,array,float,caption,placeins}


\title{The {C}-index illusion: discrimination without calibration in published survival models}

\author{\name Rafael da Silva \email rafael.dasilva@eastern.edu \\
       \addr PhD in Applied Data Science\\
       Eastern University\\
       1300 Eagle Road, Saint Davids, 19087, Pennsylvania, USA \medskip \\
       \name Danilo Alvares \email danilo.alvares@mrc-bsu.cam.ac.uk \\
       \addr MRC Biostatistics Unit\\
       University of Cambridge\\
       Forvie Site, Robinson Way, Cambridge, CB20SR, UK
       }

\def\month{MM}
\def\year{YYYY}
\def\openreview{\url{https://openreview.net/forum?id=XXXX}}

\begin{document}

\maketitle

\begin{abstract}
Recent studies have argued normatively, on synthetic data, that evaluating survival models by discrimination alone can produce misleading assessments because the concordance index ignores calibration and time-dependent accuracy. We reproduce three published survival-ML models across hard-drive failure, peer-to-peer credit, and digital-platform disengagement, validate our instrument against the anchor paper's synthetic experiment, and test five pre-registered hypotheses with Holm correction. H1--H5 are interpreted separately: H2, H3, and H5 reject under the literal protocol reading, whereas H1 and H4 do not; under a proportional sensitivity reading of H3, only H2 and H5 reject. Domain~1 D-Calibration rejects calibration conditional on non-informative censoring, but is an in-sample goodness-of-fit result. A post-hoc fixed-horizon analysis finds observed risks exceeding mean predictions by 0.0117, 0.0134, and 0.0214 at 12, 24, and 36 months. Domain~2 shows upward estimator bias when prepayment is treated as censoring rather than a competing risk; this is a methodological audit of the evaluation procedure, not a direct failure of the audited Cox predictions. Domain~3 H5 establishes higher absolute prediction error at later horizons. A post-hoc comparison with a training-set Kaplan--Meier reference yields positive Brier skill at all horizons and does not support relative deterioration; the originally planned temporal-AUC sensitivity likewise does not support temporal decline. These hypothesis-specific results identify limitations not captured by the global C-index without constituting an omnibus test. We release a reusable, pre-registered evaluation harness.
\end{abstract}


\section{Introduction}
\label{sec:introduction}

Survival models are increasingly applied outside medicine, predicting outcomes that range from system failure in reliability engineering to event-history outcomes such as purchase and divorce \citep{emmertstreib2019,wang2019}. Across these applications, model comparison and reporting rest almost entirely on a single number: the concordance index (or C-index). A recent ICML 2026 Spotlight position paper, \emph{"Stop chasing the C-index when evaluating survival analysis models"} \citep{lillelund2025b}, made this practice explicit and argued against it. Based on a structured survey of 92 survival-analysis papers published between 2023 and 2025, the authors show that the overwhelming majority report the C-index alone, i.e., a metric that captures only discrimination, the ability to correctly rank pairs of subjects by risk, while remaining blind to calibration and to time-dependent predictive accuracy. Their central claim is that evaluation metrics and modeling assumptions must be aligned on the same rung of what they call the \emph{double-helix ladder}: an evaluation metric built for independent censoring cannot be trusted to assess a model that departs from that assumption, and vice versa. Misalignment, they argue, produces model comparisons that are not merely imprecise but actively misleading.

This argument was demonstrated normatively, on synthetic data. The authors generate event times from a Weibull process with covariate-dependent hazards and induce dependent censoring through a Clayton copula, which allows them to compute \emph{oracle} metrics -- quantities that are only available because the data-generating process is known and the true event and censoring times are recorded. Under this controlled setting, they show that metric bias grows systematically as censoring departs from independence. What remains untested is whether this matters in practice: on real, published models, outside the clinical domain where survival analysis originated and where assumption-aware evaluation tools were developed. SurvBoard, a prominent multi-omics survival benchmark \citep{wissel2025}, is itself built exclusively on oncological data. Outside healthcare, the double-helix ladder has never been applied to a real, published model.

This paper takes that step. We ask a direct question: does \emph{"stop chasing the C-index"} hold up outside healthcare, on real data, in models that have already been published? To find out, we reproduce three published survival-ML models across three structurally distinct non-clinical domains -- engineering reliability, peer-to-peer credit, and digital-platform analytics -- and subject each to the evaluation ladder the anchor paper proposes. Before applying the ladder to real data, we validate our own implementation of it by reproducing the anchor's synthetic experiment inside our pipeline, ensuring that any miscalibration or bias we report downstream is attributable to the models, data, and evaluation procedures under audit, not to our own instrumentation. We then test five pre-registered hypotheses, each isolating a distinct mechanism by which C-index-only evaluation can mislead -- ranking inversion, discrimination-calibration dissociation, competing-risks blindness, feature-concentration in discrimination, and horizon-specific increases in absolute prediction error -- with Holm correction across H1--H5.

We frame this study explicitly as a first step. The anchor paper's Spotlight distinction signals that the ML community considers its normative argument important; our contribution is to evaluate that argument on real, published models, and to do so transparently enough that others can extend the test to further domains and models. We release a reusable, pre-registered evaluation harness for exactly that purpose.

Our results provide hypothesis-specific evidence that global discrimination can omit consequential aspects of performance in the evaluated models and domains. The direct ranking-inversion test does not reject. H2 identifies a discrimination--calibration dissociation conditional on non-informative censoring, H3 identifies competing-risks estimator bias under its literal rule, and H5 identifies higher absolute prediction error at later horizons without post-hoc evidence of relative deterioration. The remainder of the paper is organized as follows. Section~\ref{sec:background} reviews the evaluation tools and double-helix ladder. Section~\ref{sec:methods} details our reproduction protocol, instrument validation, and hypotheses. Section~\ref{sec:results} reports hypothesis-specific results. Section~\ref{sec:discussion} discusses their scope and limitations. Section~\ref{sec:conclusion} concludes.

\section{Background \& related work}
\label{sec:background}

\subsection{Evaluating survival models}
\label{subsec:eval-survival}

The concordance index (or C-index) is by far the most widely reported metric in survival analysis. In its most common form, also known as Harrell's C \citep{harrell1982}, it estimates the probability that, for a randomly chosen pair of subjects, the model correctly ranks the one who experiences the event first as having higher predicted risk. Uno's C \citep{uno2011} refines this estimate under different assumptions about censoring, and Antolini's time-dependent C \citep{antolini2005} extends concordance to models whose risk ranking may change over time; a recent study shows that the choice of variant is not a matter of taste: different C-index formulations can disagree enough to change which model appears superior \citep{rossi2025}. What all variants share is a structural limitation, i.e., they measure \emph{discrimination}, the ordering of risk, and nothing else. A model can achieve near-perfect discrimination while assigning probabilities that are systematically wrong, because discrimination is invariant to any monotonic transformation of the predicted risk.

\emph{Calibration} is the property the C-index cannot see. We use D-Calibration \citep{haider2020}, a formal hypothesis test that checks whether predicted survival quantiles are uniform against observed outcomes. Its interpretation is conditional on non-informative censoring. We complement it with the Integrated Brier Score under inverse-probability-of-censoring weighting (IPCW-IBS) \citep{graf1999}, a proper scoring rule that penalizes the full predicted probability, not merely its rank. Neither tool is new; both belong to a body of survival-analysis methodology that has existed, largely within Biostatistics, for over a decade. We use them here exactly as intended by their original authors -- none of the evaluation machinery in this paper is our own invention.

\subsection{The double-helix ladder}
\label{subsec:ladder}

The anchor position paper's contribution is not any single metric but a way of organizing the relationship between metrics and modeling assumptions. Right-censored survival data can be assumed to arise under increasingly demanding censoring mechanisms: censoring independent of everything (random), censoring independent of risk given covariates, or censoring that depends on the very risk being modeled (dependent censoring). The anchor paper's \emph{double-helix ladder} argues that an evaluation metric is only valid at the same rung of this hierarchy as the model it evaluates -- a metric that assumes independent censoring cannot fairly judge a model fit under dependent censoring, and using it anyway produces systematically biased comparisons.

They demonstrate this with a controlled synthetic experiment: event times generated from a Weibull distribution with covariate-dependent hazards, and censoring induced through a Clayton copula at three levels of dependence (Kendall's $\tau$ = 0.25, 0.50, 0.75). Because the data-generating process is known, they can compute \emph{oracle} metrics -- the true discrimination and proper score, unavailable in any real dataset -- and show that the bias of standard, censoring-blind metrics grows with the strength of dependence. Related work by \citet{lillelund2025a} proposes copula-adjusted, censoring-aware analogs of these metrics for settings where the dependence structure is not directly observable, at the cost of an unidentifiable sensitivity parameter that must be swept rather than estimated.

\subsection{Competing risks and informative censoring}
\label{subsec:competing-risks}

A closely related and, in applied domains, more common failure mode is treating a \emph{competing event} as if it were non-informative censoring. When a subject can experience one of several mutually exclusive events -- for instance, loan default versus early repayment -- treating the competing event as ordinary censoring biases the estimated risk of the event of interest upward: the naive, cause-blind Kaplan-Meier estimator of cumulative incidence is greater than or equal to the Aalen-Johansen estimator whenever competing events have positive hazard, and the two coincide when there are no competing events \citep{satagopan2004,coemans2022}. Fine and Gray's subdistribution hazards model \citep{fine1999} and the validation guidance of \citet{vangeloven2022}, which provides recommended tools for handling this correctly, are the established methodological references here; \citet{jeanselme2025} document the downstream cost of ignoring competing risks, including its effect on fairness. In credit risk specifically, treating prepayment as censoring rather than as a competing risk to default is a documented and long-standing concern \citep{wycinka2019,frydman2020}.

\subsection{Survival ML outside healthcare}
\label{subsec:survival-ml-outside}

Survival modeling has moved well beyond its clinical origin. Published applications now span equipment-failure prediction, credit risk in peer-to-peer lending, and user-retention modeling on digital platforms. The three baselines audited in this paper -- \citet{ahmed2024} on hard-drive failure, \citet{bonewinkel2024} on peer-to-peer credit default, and \citet{abedi2022} on user disengagement in online communities -- were selected because they satisfy four conditions simultaneously: the underlying dataset is public, a survival-ML model is fit to it, a C-index is explicitly reported, and the domain lies outside healthcare. We verified directly, by reading each paper's evaluation section, that none of the three reports any calibration metric or proper scoring rule; evaluation in all three cases rests entirely on discrimination. Notably, \citet{bonewinkel2024} observe in passing that the platform's own credit ratings ``may not be well-calibrated,'' without testing this formally, and separately note that loan prepayment could in principle be modeled as a competing risk to default -- a limitation they acknowledge but do not address. We take this as an invitation rather than an oversight: it is precisely the kind of assumption-blind evaluation the anchor paper warns against, observed in the wild.

This selection criterion -- reporting a C-index, out of any number of survival-ML papers in these domains -- is a necessary condition for the audit that follows, not a claim that the field as a whole reports only discrimination. Our sample of three is not intended to be representative of non-clinical survival ML in general; it is intended to demonstrate that the specific failure mode the anchor paper describes is not confined to synthetic data or to clinical applications, wherever it can be found.

\section{Methods}
\label{sec:methods}

\subsection{Pipeline overview}
\label{sec:pipeline}

\begin{figure}[htbp!]
\centering
\includegraphics[width=0.88\linewidth,height=0.72\textheight,keepaspectratio]{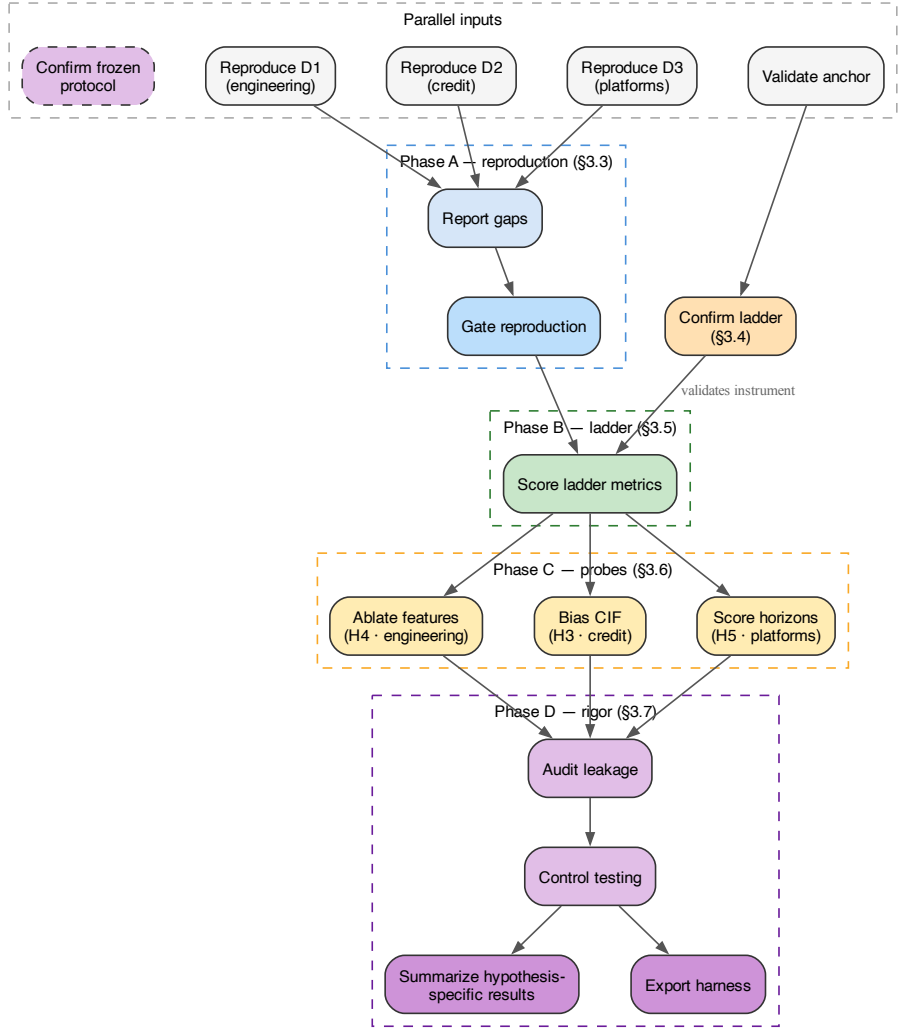}
\caption{Overview of the study pipeline. Three published baselines and the anchor's synthetic experiment enter in parallel. Phase A reproduces each baseline against its reported quantities; independently, the anchor's synthetic experiment is reproduced to validate the evaluation instrument. Phase B applies the evaluation ladder -- discrimination, calibration, and proper scoring -- to every reproduced model. Phase C applies domain-specific failure-mode probes (H3-H5). Phase D applies pre-registered rigor controls and emits the hypothesis-specific summary and reusable harness (see Subsection~\ref{sec:pipeline}).}
\label{fig:workflow}
\end{figure}

Figure~\ref{fig:workflow} summarizes the full pipeline. Three published baselines and the anchor's synthetic experiment enter in parallel. Baseline reproduction (Phase A) yields a per-domain gap report against each paper's headline quantities, gated by the reproduction criterion described in Subsection~\ref{subsec:phase-a}. Independently, we reproduce the anchor's synthetic experiment inside our own pipeline (Subsection~\ref{subsec:instrument-validation}) to validate that our implementation of the evaluation ladder behaves as the anchor's authors intended before we apply it to real data. Once a domain's reproduction clears the gate, its models feed Phase B, where we compute discrimination, calibration, and proper-scoring metrics under a fixed evaluation lens (Subsection~\ref{subsec:phase-b}). Phase C applies domain-specific failure-mode probes -- competing-risks bias in the credit domain, feature ablation in the engineering domain, and horizon-specific scoring in the platform domain (Subsection~\ref{subsec:phase-c}). Finally, Phase D wraps the whole pipeline in pre-registration, leakage auditing, and multiple-testing control (Subsection~\ref{subsec:phase-d}), producing the Holm-adjusted hypothesis-specific results reported in Section~\ref{sec:results}.

\subsection{Baseline selection}
\label{sec:baseline-selection}

We selected three published survival-ML papers under four simultaneous criteria: the underlying dataset must be public, the model must be a survival-analysis method, a concordance index must be explicitly reported, and the domain must lie outside healthcare. Table~\ref{tab:domains} summarizes the three domains that satisfied all four criteria: engineering reliability (hard-drive failure), structured finance (peer-to-peer credit default), and digital-platform analytics (user disengagement).

\begin{table}[!htbp]
\centering
\caption{The three domains audited in this study. All three datasets are public; selection criteria are described in Subsection~\ref{sec:baseline-selection}.}
\label{tab:domains}
{\footnotesize
\resizebox{\linewidth}{!}{%
\begin{tabular}{llllll}
\toprule
\textbf{Domain} & \textbf{Baseline (Author, Year)} & \textbf{Dataset} & \textbf{Event} & \textbf{Reported C-index} & \textbf{Public} \\
\midrule
D1: Engineering Reliability & Ahmed \& Green (2024) & Backblaze Drive Stats (SMART) & Hard-drive failure & 0.958 & Yes \\
D2: Structured Finance & Bone-Winkel \& Reichenbach (2024) & Bondora P2P Loans & Loan default & --$^{\dagger}$ & Yes \\
D3: Digital Platforms & Abedi Firouzjaei (2022) & Stack Exchange Data Dump & User disengagement & 0.61--0.76 (range across 18 cells) & Yes \\
\bottomrule
\multicolumn{6}{l}{\footnotesize $^{\dagger}$The baseline's headline evaluation is rating-stratified rather than a single pooled C-index;} \\
\multicolumn{6}{l}{\footnotesize per-model values (0.659 linear, 0.674 boosted) appear in their \S4.1 and are reported in Table~\ref{tab:repro-headline}.}
\end{tabular}
}
}
\end{table}

\subsection{Phase A: baseline reproduction}
\label{subsec:phase-a}

For each baseline, we replicated the reported feature engineering, model class, train/test split, and hyperparameters as closely as the published text allowed, and refit the model on the corresponding public dataset. We treat the resulting gap between our reproduced quantities and the paper's reported values as evidence in its own right: a headline C-index that can only be recovered under specific, undocumented preprocessing choices is itself a finding about the fragility of the original result, not merely a nuisance to be minimized.

Because none of the three papers pre-specifies a numerical reproduction tolerance, we did not pre-register one either. We report this honestly: the admission criterion below was fixed only after observing that all three domains cleared it comfortably, and we verified that it governs baseline \emph{admission} into the study, not the outcome of any hypothesis test -- no conclusion in Section~\ref{sec:results} depends on where this threshold is drawn. A reproduced quantity within 0.01 of its reported value is classified as a \emph{strict} match; within $(0.01, 0.03]$ as \emph{approximate}; beyond 0.03 as a \emph{reproduction gap}, reported and retained rather than discarded. No domain was excluded from the study on reproduction grounds.

Full preprocessing details, including every point at which the published text under-specifies a choice we had to make, are documented per domain in Appendix~\ref{app:reproduction}, together with the exact wording we use in Section~\ref{sec:discussion} to describe each domain's residual gap.

\subsection{Instrument validation: the anchor harness}
\label{subsec:instrument-validation}

Before applying the evaluation ladder to any real model, we validated our implementation of it against the one setting where ground truth is knowable: the anchor paper's own synthetic experiment. We ported the authors' published data-generating process and evaluation code into our pipeline and reproduced their Table 2 (oracle discrimination and proper score under five censoring regimes, including three levels of Clayton-copula dependence) and their Figure 5 (the bias of censoring-blind metrics relative to the oracle, under the same five regimes) with 100 random seeds. The maximum absolute gap on any oracle quantity is 0.0004 for discrimination and 0.0005 for proper score; Figure 5 bias metrics match at the level of $10^{-7}$ for Harrell C and uncensored IBS, and $10^{-4}$ for IPCW quantities.

This validation establishes fidelity of \emph{integration}, not independent correctness: we did not re-implement the anchor's metrics from scratch, so a latent error in the authors' own code would be inherited rather than caught. What it does establish is that our orchestration -- our data loaders, our environment, our handling of the anchor's data structures -- does not introduce error at the boundary where their code meets our pipeline. Any discrepancy we report in Section~\ref{sec:results} is therefore attributable to the real-world models, data, and evaluation procedures under audit, not to a mismatch between our harness and the instrument it wraps. We apply these validated copula-aware evaluation tools to real data in Subsection~\ref{subsec:sensitivity}, where we sweep the unidentifiable dependence parameter across the range the anchor paper tested.

\begin{table}[!htbp]
\centering
\caption{Reproduction of the anchor paper's Table 2 (oracle discrimination and proper score under five censoring regimes) inside our pipeline. Oracle metrics use the true, simulation-known event times and are only available in this synthetic setting. See Subsection~\ref{subsec:instrument-validation}.}
\label{tab:anchor-oracle}
{\footnotesize
\resizebox{\linewidth}{!}{%
\begin{threeparttable}
\begin{tabular}{lcccccc}
\toprule
\textbf{Censoring Scenario} & \textbf{$C_{\mathrm{oracle}}$ (ours)} & \textbf{$C_{\mathrm{oracle}}$ (paper)} & \textbf{Gap (C)} & \textbf{IBS$_{\mathrm{oracle}}$ (ours)} & \textbf{IBS$_{\mathrm{oracle}}$ (paper)} & \textbf{Gap (IBS)} \\
\midrule
Random & 0.6344 & 0.6340 & $+0.0004$ & 0.0900 & 0.0900 & $+0.0000$ \\
Independent & 0.6341 & 0.6340 & $+0.0001$ & 0.0837 & 0.0840 & $-0.0003$ \\
Dependent ($\tau{=}0.25$) & 0.6281 & 0.6280 & $+0.0001$ & 0.1322 & 0.1320 & $+0.0002$ \\
Dependent ($\tau{=}0.50$) & 0.6179 & 0.6180 & $-0.0001$ & 0.1995 & 0.1990 & $+0.0005$ \\
Dependent ($\tau{=}0.75$) & 0.6092 & 0.6090 & $+0.0002$ & 0.2453 & 0.2450 & $+0.0003$ \\
\bottomrule
\end{tabular}
\begin{tablenotes}
    \small
    \item \textit{Note.} 100 random seeds. Maximum absolute gap: 0.0004 (discrimination), 0.0005 (proper score).
    \end{tablenotes}
\end{threeparttable}
}
}
\end{table}

\subsection{Phase B: the evaluation ladder}
\label{subsec:phase-b}

With each baseline reproduced and the instrument validated, we hold every model fixed and vary only the evaluation lens. For each reproduced model we compute: discrimination, via Harrell's C and Uno's IPCW-adjusted C; calibration, via the D-Calibration hypothesis test ($\chi^2$ statistic over ten quantile bins); and a proper scoring rule, via the IPCW-weighted Integrated Brier Score. All three metrics are computed with a single backend (\texttt{scikit-survival}) across all domains, chosen for consistency with the instrument validated in Subsection~\ref{subsec:instrument-validation}.

\subsection{Phase C: failure-mode probes}
\label{subsec:phase-c}

Three domain-specific probes target the mechanisms our pre-registered hypotheses isolate. In the credit domain, we compute the cumulative incidence of default at a twelve-month horizon under both the naive, competing-risks-blind estimator and the Aalen-Johansen estimator, stratified by credit rating, with bootstrap confidence intervals ($B = 1000$) (H3). H3 is an audit of the cumulative-incidence estimator used to interpret default risk under competing events; it is not a direct test of the Cox models' predicted survival curves. In the engineering domain, we conduct a broad leave-one-out ablation survey over SMART attributes, testing -- for every feature individually -- whether its removal produces a discrimination drop large enough, and consistent enough across paired bootstrap resamples ($B = 1000$), to indicate that the model's discrimination is concentrated rather than distributed (H4). This is deliberately an existential test: we do not commit in advance to any specific feature being responsible, only to searching broadly and reporting whatever the search finds, including a null result. In the platform domain, we compute the pointwise IPCW Brier score at twelve-, twenty-four-, and thirty-six-month scoring horizons for each random survival forest, testing whether absolute prediction error increases with horizon even while the model's global C-index remains within the range the original authors reported (H5).

\subsection{Phase D: rigor controls}
\label{subsec:phase-d}

Every transformation used in reproduction and evaluation is fit exclusively on training data; a leakage audit confirms this across all eight checkpoints. Because the hypotheses are not independent, we apply a two-level Holm--Bonferroni correction: an inner level collapses hypotheses with multiple sub-tests to one decision $p$-value, and an outer level corrects across H1--H5. We report the five Holm-adjusted results separately. The pre-registered count rule is retained in Appendix~\ref{app:protocol} as a historical record, but after review we do not treat it as a calibrated omnibus test or assign it a global inferential verdict.

\subsection{Post-hoc peer-review-response analyses}
\label{subsec:posthoc-review1}
Two analyses were added post hoc in response to peer review and do not replace the pre-registered tests. For H2, we assess the frozen Domain~1 \texttt{cox\_full} model at 12, 24, and 36 months (365.25, 730.5, and 1,095.75 days). We report mean predicted risk, Kaplan--Meier observed incidence, observed-minus-predicted difference, and ICI using five prediction-defined groups. This remains an \texttt{in\_sample\_gof} assessment and is conditional on non-informative censoring. A planned out-of-sample sensitivity was gated before estimation and not run because the train-only Cox refit did not converge under the frozen choices, including robustness attempts with up to 2,000 optimizer steps.

For H5, we compare the frozen \texttt{rsf\_content} model with a constant marginal Kaplan--Meier reference estimated on training data. At each horizon, $\mathrm{BSS}(t)=1-\mathrm{Brier}_{\mathrm{model}}(t)/\mathrm{Brier}_{\mathrm{reference}}(t)$. Paired bootstrap intervals resample identical test subjects for model and reference ($B=1{,}000$, seed 42). This reference-relative analysis receives no new reject/non-reject label.

\subsection{Pre-registered hypotheses}
\label{subsec:hypotheses}

Table~\ref{tab:hypotheses} summarizes H1--H5. Their exact frozen nulls, alternatives, statistics, and decision rules are preserved in Appendix~\ref{app:protocol}.

\begin{table}[!htbp]
\centering
\caption{The five pre-registered hypotheses. Holm-adjusted results are interpreted separately; no omnibus inference is assigned to their count.}
\label{tab:hypotheses}
{\footnotesize
\begin{threeparttable}
\begin{tabular}{@{}c>{\raggedright\arraybackslash}p{2.4cm}c>{\raggedright\arraybackslash}p{3.2cm}>{\raggedright\arraybackslash}p{3.4cm}@{}}
\toprule
\textbf{\#} & \textbf{Hypothesis} & \textbf{Domain} & \textbf{Test Statistic} & \textbf{Rejection Threshold} \\
\midrule
H1 & Ranking inversion & All & $\tau_K$ (C-index vs. IPCW-IBS rank) & $\tau_K\leq0.5$ and bootstrap $p<0.05$ \\
H2 & Discrimination--calibration dissociation & D1 & D-Calibration $p$-value & $C_H\geq0.90$ and $p<0.05$ \\
H3 & Competing-risks bias & D2 & $|F_{\mathrm{naive}}-F_{\mathrm{AJ}}|$ at 12mo & $>0.02$, CI excludes 0, and $\geq3$ strata$^{\ddagger}$ \\
H4 & Feature-concentration ablation & D1 & $\exists$ ablation set with $\Delta C_H$ & $\geq0.03$ and CI non-overlapping \\
H5 & Increase in absolute prediction error & D3 & Brier(36mo)$-$Brier(12mo) & Monotonic increase, $>2$ combined SE, and C in reported band \\
\bottomrule
\end{tabular}
\begin{tablenotes}
\footnotesize
\item $^{\ddagger}$The literal frozen rule uses an absolute count of three; the proportional sensitivity requires five of seven observed strata.
\end{tablenotes}
\end{threeparttable}
}
\end{table}

\section{Results}
\label{sec:results}

\subsection{Baseline reproduction}
\label{subsec:results-repro}

Table~\ref{tab:repro-headline} reports the headline reproduction gaps for all three domains. Domain 1 (Backblaze) reproduces to within 0.0015 of the reported Harrell C-index (0.9595 vs.\ 0.958) -- a strict match by the criterion of Subsection~\ref{subsec:phase-a}. This required identifying the exact cohort the original authors used: not the full population of drives, but healthy drives filtered to a calendar age above seven years, combined with \emph{all} failed drives regardless of age. Our resulting counts (12{,}815 healthy, 5{,}089 failed) closely track the paper's reported figures (12{,}993 and 4{,}889). We treat this as a resolved reproduction finding rather than a residual limitation: an earlier iteration of this pipeline, fit on the full unfiltered population, reproduced a C-index of 0.981 -- a gap of $+0.023$ that, as we discuss in Subsection~\ref{subsec:h4}, later proved diagnostic in its own right.

Domain 2 (Bondora) reproduces the linear Cox model to a strict match (0.6559 vs.\ 0.659) and the gradient-boosted variant to an approximate match (0.6504 vs.\ 0.674); the residual gap is isolated to boosting hyperparameters the original paper does not fully specify. Domain 3 (Stack Exchange) reproduces all eighteen reported cells of the original Table 8 to a mean absolute gap of 0.0035, following alignment to the author's PySurvival backend. Full per-domain preprocessing decisions, including every point where the published text required interpretation, are documented in Appendix~\ref{app:reproduction}.

\begin{table}[!htbp]
\centering
\caption{Reproduction gaps for each baseline's headline discrimination quantity. Full per-domain detail is in Appendix~\ref{app:reproduction}; the complete Domain~3 table is in Appendix~\ref{app:fullrepro}.}
\label{tab:repro-headline}
{\footnotesize
\resizebox{\linewidth}{!}{%
\begin{threeparttable}
\begin{tabular}{llcccc}
\toprule
\textbf{Domain} & \textbf{Model} & \textbf{C (reported)} & \textbf{C (reproduced)} & \textbf{Gap} & \textbf{Tier} \\
\midrule
D1 & Cox & 0.9580 & 0.9595 & $+0.0015$ & Strict \\
D2 & Cox (linear) & 0.6590 & 0.6559 & $-0.0031$ & Strict \\
D2 & Cox (boosted) & 0.6740 & 0.6504 & $-0.0236$ & Approximate \\
D3 & RSF (18-cell mean) & 0.61--0.76 & 0.61--0.76 & n.a. & Strict \\
\bottomrule
\end{tabular}%
\begin{tablenotes}
    \footnotesize
    \item \textit{Note.} Tiers per the reproduction criterion of Subsection~\ref{subsec:phase-a}: strict $\leq 0.01$, approximate $(0.01, 0.03]$. Domain~3 reports the published and reproduced cell-wise bands; mean absolute gap across all 18 cells is $0.0035$ (strict). D1's earlier iteration (unfiltered cohort) reproduced $C = 0.981$; see Subsection~\ref{subsec:results-repro} and Subsection~\ref{subsec:repro-revealed}.
    \end{tablenotes}
\end{threeparttable}
}
}
\end{table}

\begin{table}[!htbp]
\centering
\caption{Complete Domain 2 reproduction quantities.}
\label{tab:d02full}
{\footnotesize
\resizebox{\linewidth}{!}{%
\begin{threeparttable}
\begin{tabular}{lp{2.4cm}p{3.0cm}cc}
\toprule
\textbf{Quantity} & \textbf{Reported} & \textbf{Source} & \textbf{Ours} & \textbf{Gap} \\
\midrule
Train/test split date & 2020-01-01 & Baseline \S3.3 & 2020-01-01 & n.a. \\
Test-set size & n.r. & Baseline \S3.3 & 6{,}457 & n.a. \\
Rating strata & 7 & Baseline Table 1 / \S3.5 & 7 & $+0.0000$ \\
Harrell C (linear Cox, test) & 0.6590 & Baseline \S4.1 & 0.6559 & $-0.0031$ \\
Harrell C (boosted Cox, test) & 0.6740 & Baseline \S4.1 & 0.6504 & $-0.0236$ \\
AA default rate, Bondora (completed) & 0.1726 & Baseline Table 1 / footnote 14 & 0.2023 & $+0.0297$ \\
AA default rate, linear Cox (completed) & n.r. & This work & 0.1723 & n.a. \\
AA default rate, boosted Cox (completed) & 0.1445 & Baseline Table 1 / footnote 14 & 0.1486 & $+0.0041$ \\
$N$ loans in boosted AA bucket & 713 & Baseline Table 1 & 467 & $-246$ \\
$N$ loans in linear AA bucket & 858 & Baseline Appendix D & 791 & $-67$ \\
AA default rate, Bondora (KM at term) & 0.1726 & Baseline \S3.6 KM & 0.1706 & $-0.0020$ \\
AA default rate, Bondora (empirical) & 0.1726 & Baseline Table 1 & 0.2102 & $+0.0376$ \\
AA default rate, linear Cox (KM at term) & n.r. & This work & 0.1593 & n.a. \\
AA default rate, boosted Cox (KM at term) & 0.1445 & Baseline Table 1 / \S3.6 KM & 0.1263 & $-0.0182$ \\
IRR, Bondora AA & $-0.0320$ & Baseline Table 1 & n.a. & n.a. \\
IRR, boosted AA & 0.1563 & Baseline Table 1 & n.a. & n.a. \\
\bottomrule
\end{tabular}
\begin{tablenotes}
    \footnotesize
    \item \textit{Note.} n.r.\ = not reported in the baseline; n.a.\ = not applicable (no baseline value to compare, or not computed here). IRR (internal rate of return) requires repayment microdata unavailable in the public dump; linear-Cox AA rates are not published in the baseline main text and are reported from this reproduction only.
    \end{tablenotes}
\end{threeparttable}
}
}
\end{table}

\begin{table}[!htbp]
\centering
\caption{Complete reproduction of all 18 cells from the original paper's Table 8 (Stack Exchange). Mean absolute gap $= 0.0035$; 12 of 18 cells within the strict tier ($\leq 0.01$). See Subsection~\ref{subsec:results-repro}.}
\label{tab:d03full}
{\footnotesize
\resizebox{\linewidth}{!}{%
\begin{tabular}{llcccc}
\toprule
\textbf{Community} & \textbf{Feature set} & \textbf{$\theta$} & \textbf{C (paper)} & \textbf{C (ours)} & \textbf{Gap} \\
\midrule
Politics & Behavioural & 24 & 0.7500 & 0.7507 & $+0.0007$ \\
 &  & 36 & 0.7600 & 0.7615 & $+0.0015$ \\
 & Content & 24 & 0.6800 & 0.6746 & $-0.0054$ \\
 &  & 36 & 0.6800 & 0.6824 & $+0.0024$ \\
 & Combined & 24 & 0.7500 & 0.7437 & $-0.0063$ \\
 &  & 36 & 0.7600 & 0.7552 & $-0.0048$ \\
Data Science & Behavioural & 24 & 0.6600 & 0.6580 & $-0.0020$ \\
 &  & 36 & 0.6600 & 0.6611 & $+0.0011$ \\
 & Content & 24 & 0.6100 & 0.6070 & $-0.0030$ \\
 &  & 36 & 0.6300 & 0.6300 & $+0.0000$ \\
 & Combined & 24 & 0.6800 & 0.6760 & $-0.0040$ \\
 &  & 36 & 0.7000 & 0.6958 & $-0.0042$ \\
Computer Science & Behavioural & 24 & 0.6800 & 0.6784 & $-0.0016$ \\
 &  & 36 & 0.6800 & 0.6782 & $-0.0018$ \\
 & Content & 24 & 0.6200 & 0.6183 & $-0.0017$ \\
 &  & 36 & 0.6300 & 0.6238 & $-0.0062$ \\
 & Combined & 24 & 0.6900 & 0.6839 & $-0.0061$ \\
 &  & 36 & 0.6800 & 0.6892 & $+0.0092$ \\
band\_coverage & -- & -- & 18 & 12 & $-6$ \\
\bottomrule
\end{tabular}
}
}
\end{table}

\subsection{Instrument validation}
\label{subsec:results-instrument}

As described in Subsection~\ref{subsec:instrument-validation}, our implementation of the evaluation ladder reproduces the anchor's synthetic Table 2 and Figure 5 to within 0.0004--0.0005 on oracle quantities and $10^{-4}$--$10^{-7}$ on bias metrics. We treat this as established and do not repeat the figures here; every result below is computed with the same validated instrument.

\subsection{H1 -- Ranking inversion}
\label{subsec:h1}

Table~\ref{tab:h1} reports Kendall's $\tau$ between the C-index ranking and the IPCW-IBS ranking within each domain. In Domains 1 and 3, the two metrics agree completely ($\tau = 1.000$). In Domain 2, they disagree completely ($\tau = -1.000$): the classical Cox model has the higher C-index (0.6559 vs.\ 0.6504) but the \emph{worse} proper score (IBS 0.5473 vs.\ 0.4806) -- the model a practitioner would prefer under discrimination alone is the model a practitioner would reject under a proper scoring rule. Under the pre-registered decision rule of C00.1, however, this qualitative inversion does not reach formal significance: the bootstrap test has no power to distinguish a genuine inversion from noise when only two models are being compared, and the resulting confidence interval necessarily spans the full range of $\tau$. The Holm-adjusted $p$-value across all three domains is 0.1558; H1 does not reject.

\begin{table}[!htbp]
\centering
\caption{Kendall's $\tau$ between the C-index ranking and the IPCW-Integrated-Brier-Score ranking, within each domain. $\tau = -1$ in Domain 2 indicates complete rank inversion between the two metrics; the wide confidence interval reflects the two-model comparison's lack of statistical power (Subsection~\ref{subsec:limitations}). See Figure~\ref{fig:h1}.}
\label{tab:h1}
\begin{tabular}{lccc}
\toprule
\textbf{Domain} & \textbf{$\tau_K$} & \textbf{Bootstrap 95\% CI} & \textbf{Reject (C00.1)} \\
\midrule
D1 & 1.000 & $[1.000, 1.000]$ & No \\
D2 & $-1.000$ & $[-1.000, 1.000]$ & No \\
D3 & 1.000 & $[0.333, 1.000]$ & No \\
\bottomrule
\end{tabular}
\end{table}

\begin{figure}[!htbp]
\centering
\includegraphics[width=0.88\linewidth,height=0.52\textheight,keepaspectratio]{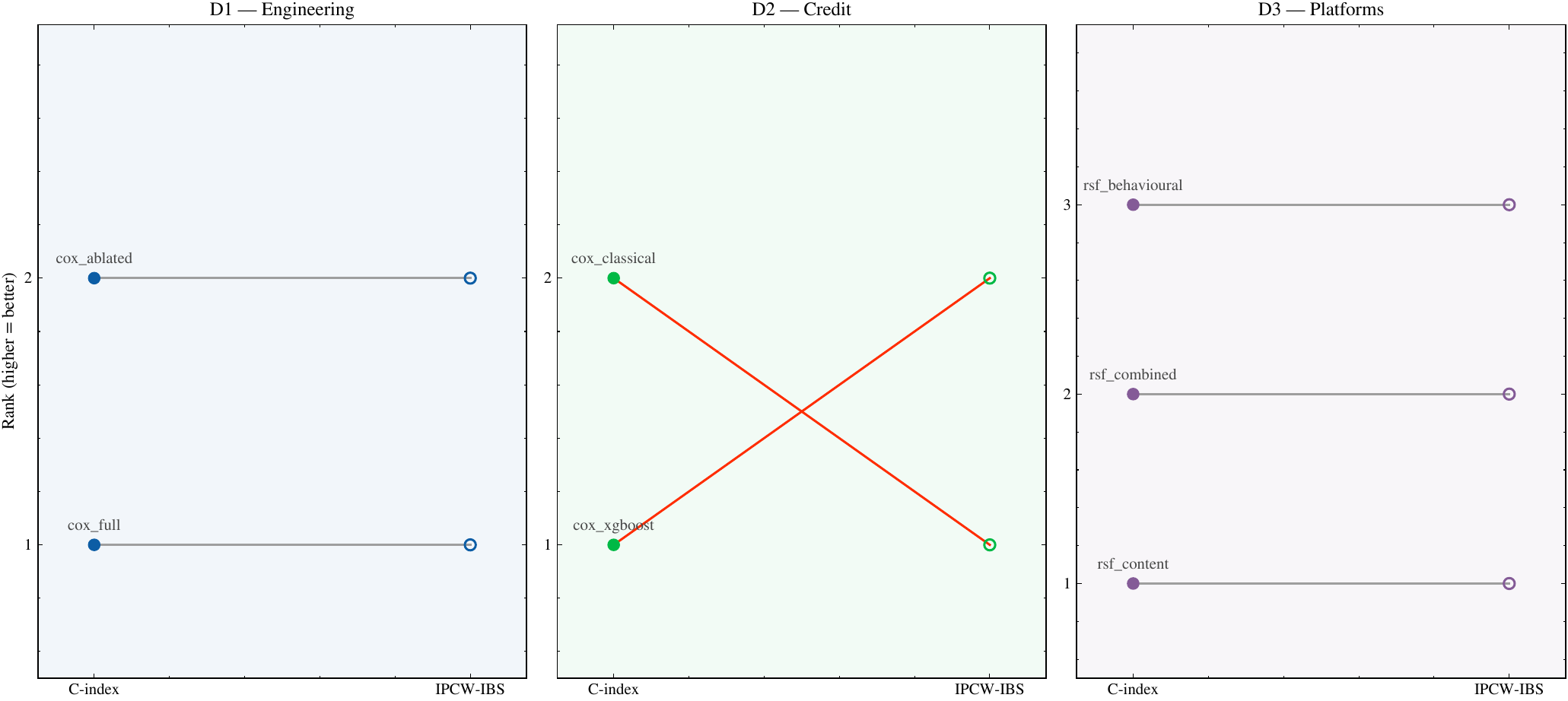}
\caption{Discrimination (C-index) versus proper-score (IPCW-IBS) rankings, by model and domain. Points are ranked (not raw-value) positions; a crossing connector indicates a rank inversion between the two metrics. Domain 2 (red connector) shows a complete inversion -- the model with the better C-index has the worse proper score -- though this does not reach formal significance (Table~\ref{tab:h1}, Subsection~\ref{subsec:limitations}).}
\label{fig:h1}
\end{figure}

\paragraph{What this means.}
H1 does not reject. Within the evaluated models and domains, discrimination and proper-score rankings agree except in Domain~2, where only two models are available and the test lacks power to distinguish the observed inversion from noise. H2--H5 examine other dimensions not summarized by the global C-index.

\FloatBarrier
\subsection{H2 -- Discrimination--calibration dissociation}
\label{subsec:h2}

Table~\ref{tab:dcal} reports D-Calibration for all seven models. Conditional on the non-informative-censoring assumption required by the procedure, every model rejects the null of calibration. H2 pre-registers Domain~1 \texttt{cox\_full}: it reproduces the published discrimination ($C=0.9595$ versus 0.958) and rejects D-Calibration at $p=2.60\times10^{-136}$; its Holm-adjusted $p$-value is $1.30\times10^{-135}$. This is an \texttt{in\_sample\_gof} result, not out-of-sample predictive validation, and the small $p$-value does not establish robustness to informative censoring.

\begin{table}[!htbp]
\centering
\caption{D-Calibration results for the reproduced models. Interpretation is conditional on non-informative censoring. The bolded Domain~1 result is an in-sample goodness-of-fit assessment.}
\label{tab:dcal}
{\footnotesize
\begin{threeparttable}
\begin{tabular}{llccccc}
\toprule
\textbf{Domain} & \textbf{Model} & \textbf{C-index} & \textbf{D-Cal $p$} & \textbf{ECE (pp)} & \textbf{Max bin (pp)} & \textbf{Reject} \\
\midrule
D1 & \textbf{cox\_full} & 0.9595 & $2.60\times10^{-136}$ & 1.78 & 2.79 & Yes \\
D1 & cox\_ablated & 0.9617 & $1.06\times10^{-33}$ & 0.86 & 1.60 & Yes \\
D2 & cox\_classical & 0.6559 & $7.03\times10^{-108}$ & 1.92 & 6.50 & Yes \\
D2 & cox\_xgboost & 0.6504 & $9.85\times10^{-161}$ & 3.26 & 9.23 & Yes \\
D3 & rsf\_behavioural & 0.7789 & $1.78\times10^{-27}$ & 1.31 & 5.38 & Yes \\
D3 & rsf\_content & 0.6982 & $5.91\times10^{-42}$ & 1.35 & 7.24 & Yes \\
D3 & rsf\_combined & 0.7739 & $6.14\times10^{-44}$ & 1.65 & 6.55 & Yes \\
\bottomrule
\end{tabular}
\begin{tablenotes}\footnotesize
\item H2 formally targets \texttt{cox\_full}; the all-model pattern is descriptive and post-hoc ECE values do not replace the pre-registered test.
\end{tablenotes}
\end{threeparttable}
}
\end{table}

\begin{figure}[!htbp]
\centering
\includegraphics[width=0.82\linewidth,height=0.48\textheight,keepaspectratio]{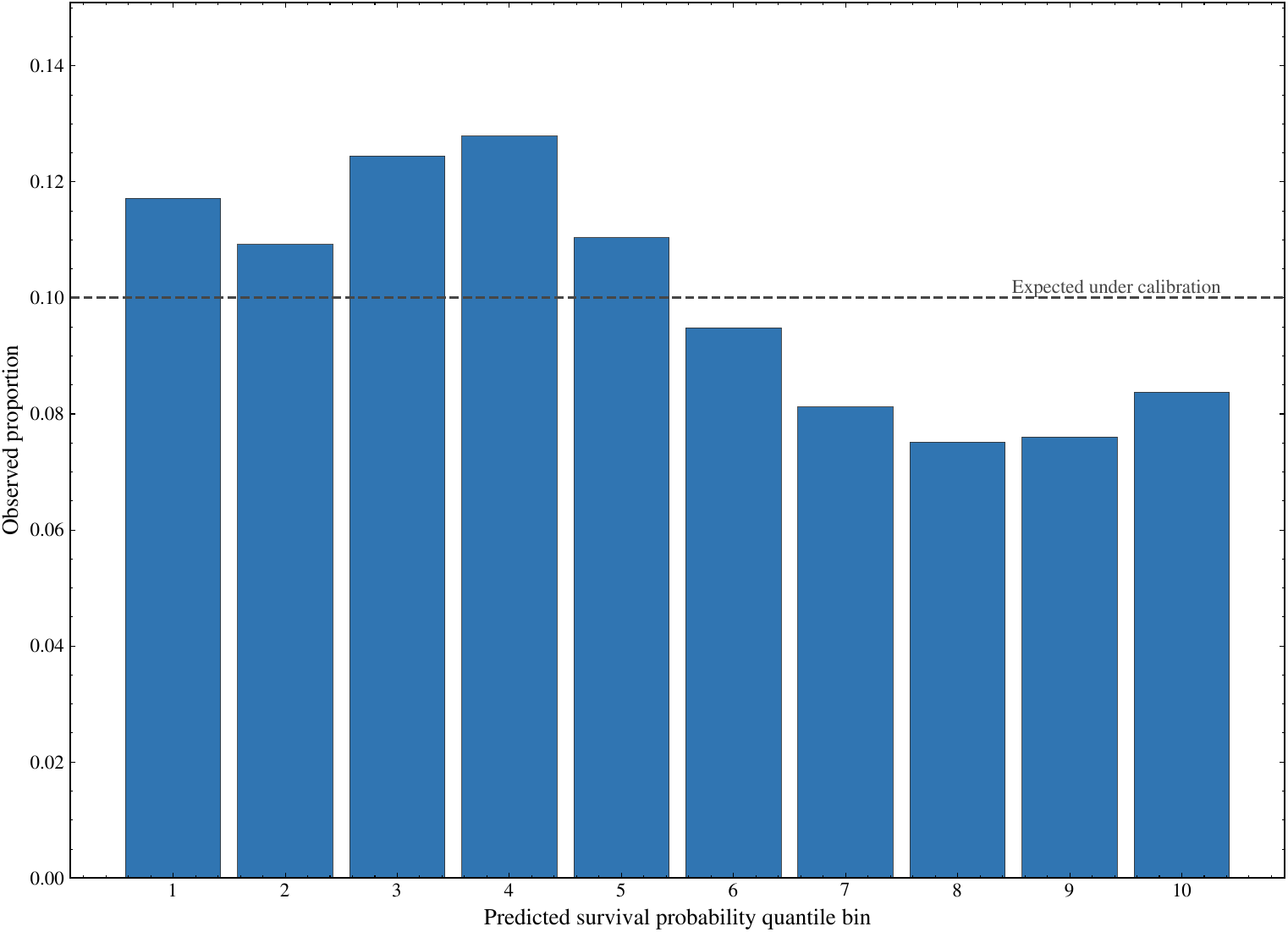}
\caption{D-Calibration histogram for Domain~1 \texttt{cox\_full}. Under the non-informative-censoring assumption, this in-sample goodness-of-fit test rejects calibration ($p=2.60\times10^{-136}$).}
\label{fig:dcal}
\end{figure}

\begin{table}[!htbp]
\centering
\caption{Post-hoc fixed-horizon calibration for Domain~1 \texttt{cox\_full} ($B=1{,}000$). Observed incidence is Kaplan--Meier; no censoring occurred before these horizons.}
\label{tab:h2fixed}
\begin{tabular}{lcccccc}
\toprule
\textbf{Months} & \textbf{Predicted} & \textbf{Observed} & \textbf{Obs.--pred.} & \textbf{ICI} & \textbf{At risk} & \textbf{Events} \\
\midrule
12 & 0.0313 & 0.0430 & 0.0117 & 0.0177 & 17{,}060 & 766 \\
24 & 0.0829 & 0.0963 & 0.0134 & 0.0207 & 16{,}110 & 1{,}716 \\
36 & 0.1241 & 0.1455 & 0.0214 & 0.0253 & 15{,}232 & 2{,}594 \\
\bottomrule
\end{tabular}
\end{table}

\begin{figure}[!htbp]
\centering
\includegraphics[width=0.95\linewidth,height=0.48\textheight,keepaspectratio]{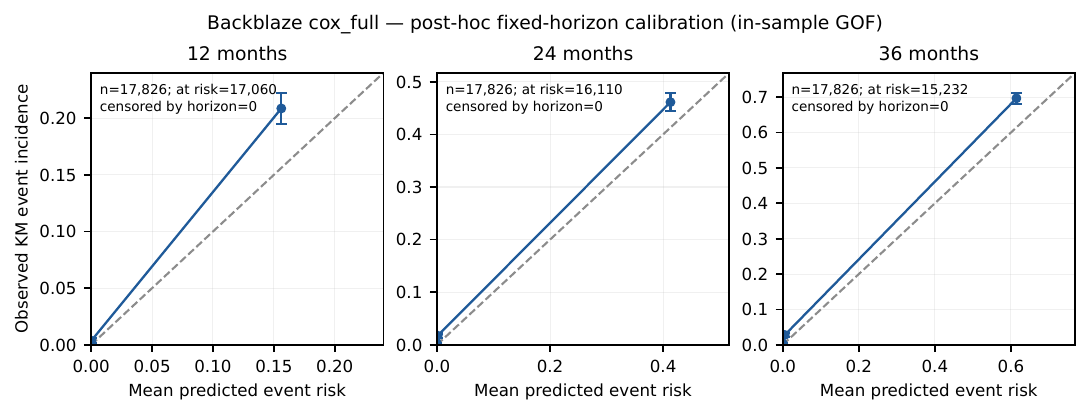}
\caption{Post-hoc fixed-horizon calibration at 12, 24, and 36 months. Points show five prediction-defined groups; the diagonal denotes perfect calibration. This is an \texttt{in\_sample\_gof} analysis conditional on non-informative censoring. There was zero censoring before each displayed horizon.}
\label{fig:h2fixed}
\end{figure}

Observed incidence exceeds mean predicted risk at each horizon, and ICI increases from 0.0177 to 0.0253. These post-hoc results improve horizon-specific interpretation but do not replace D-Calibration or address informative censoring. The OOS gate failed because the train-only Cox refit did not converge under the frozen choices, so no OOS estimate was produced.

\FloatBarrier
\subsection{H3 -- Competing-risks bias}
\label{subsec:h3}

At a twelve-month horizon, the naive cumulative incidence of default (0.2218) exceeds the Aalen-Johansen estimate (0.1999) by 0.0218, with a bootstrap confidence interval of $[0.0201, 0.0238]$ that excludes zero. Table~\ref{tab:h3} breaks this down by credit rating: the point estimate remains below the pre-registered support threshold for the safer ratings (AA, A, B, C) and grows to 2.1, 3.0, and 3.9 percentage points for the three riskiest strata (D, E, F respectively). Three of the seven observed Bondora ratings meet the support threshold -- satisfying the absolute count $\geq 3$ required by the freeze (C00 text: ``$\geq 3$ of 5,'' written when five strata were expected; see Appendix~\ref{app:protocol}). The Holm-adjusted $p$-value is 0.0448; H3 rejects, though by a narrow margin relative to the pre-registered threshold of 0.02.

\begin{table}[!htbp]
\centering
\caption{Bias in twelve-month cumulative incidence of default (naive estimator minus Aalen-Johansen), by Bondora credit rating. Bias grows monotonically with credit risk; only the three riskiest strata (D, E, F) meet the absolute support count $\geq 3$ (C00 freeze text expected five ratings; seven are observed). See Figure~\ref{fig:h3}.}
\label{tab:h3}
\begin{tabular}{lcccc}
\toprule
\textbf{Rating} & \textbf{$n$} & \textbf{$\Delta$ (naive $-$ AJ)} & \textbf{95\% CI} & \textbf{Supports H3} \\
\midrule
AA & 176 & 0.0016 & $[0.0004, 0.0037]$ & No \\
A & 114 & 0.0114 & $[0.0045, 0.0202]$ & No \\
B & 636 & 0.0078 & $[0.0052, 0.0110]$ & No \\
C & 1{,}373 & 0.0158 & $[0.0131, 0.0194]$ & No \\
D & 1{,}785 & 0.0210 & $[0.0182, 0.0243]$ & Yes \\
E & 2{,}031 & 0.0300 & $[0.0264, 0.0346]$ & Yes \\
F & 342 & 0.0390 & $[0.0258, 0.0530]$ & Yes \\
\bottomrule
\end{tabular}
\end{table}

\begin{center}
\includegraphics[width=0.85\linewidth,height=0.48\textheight,keepaspectratio]{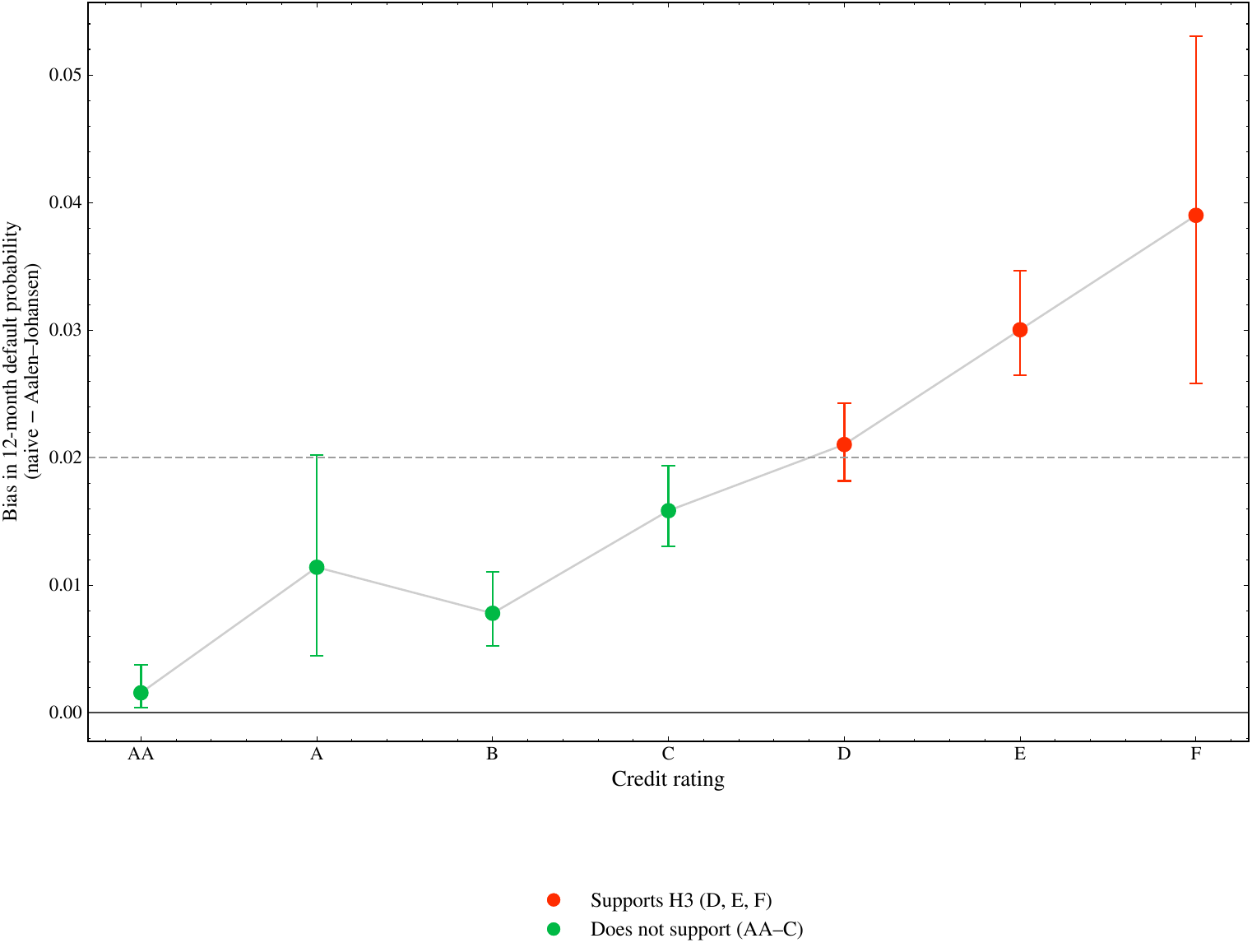}
\captionof{figure}{Bias in twelve-month cumulative incidence of default, by credit rating (safest to riskiest, left to right), with 95\% bootstrap confidence intervals. Bias grows monotonically with risk; only the three riskiest strata (red) exceed the pre-registered support threshold (dashed line).}
\label{fig:h3}
\end{center}

\paragraph{What this means.}
H3 evaluates a known competing-risks estimation issue in the Bondora data, rather than directly testing the predictions of the audited Cox models. Treating prepayment as ordinary censoring produces a higher estimated cumulative incidence of default than the Aalen-Johansen estimator, with an overall difference of 0.0218 at twelve months and larger discrepancies in the highest-risk rating strata. The contribution is to quantify the practical magnitude and heterogeneity of this established estimator bias in a published non-clinical evaluation setting. These results do not establish that the Cox models themselves are defective or miscalibrated; they show that the choice of evaluation estimator affects the absolute risk quantities used to assess and interpret survival-model results. H3 is therefore a methodological audit complementary to the direct model assessments, not an additional demonstration of model failure. We note two sources of narrowness. First, the observed bias (0.0218) exceeds the pre-registered threshold (0.02) by only 0.0018. Second, the freeze text specifies ``$\geq 3$ of 5 rating strata,'' written when five strata were expected; Bondora's data contain seven (AA--F). We applied the absolute count ($\geq 3$) as frozen, but a fractional reading ($\geq 60\%$, i.e.\ $\geq 5$ of 7) is equally defensible -- and under it H3 would not reject. The number of strata should have been verified before the protocol freeze; it was not, and we disclose rather than silently correct this. We return to both readings in Subsection~\ref{subsec:limitations}.

\FloatBarrier
\subsection{H4 -- Broad feature ablation}
\label{subsec:h4}

Across the full leave-one-out ablation survey of SMART attributes in Domain 1, no single feature's removal produces a discrimination drop of 0.03 or more with a non-overlapping bootstrap confidence interval. The protocol reject is \texttt{False}; H4 does not reject.

We additionally tested three theoretically-motivated feature clusters under the same paired-bootstrap rule ($B = 1000$, $\Delta C \geq 0.03$ and non-overlapping CIs), removing each cluster jointly from the frozen \texttt{cox\_full} fit ($C = 0.9595$). Cluster~A (age/usage proxies: SMART~9, 240, 241) drops discrimination to $C = 0.8676$ ($\Delta C = 0.0920$; bootstrap 95\% CI for $\Delta$ $[0.0842, 0.1004]$), clearing the threshold. Cluster~B (degradation trio: SMART~5, 197, 198) does not ($\Delta C = -0.0022$; CIs overlap). Cluster~C (A$\cup$B) also clears ($\Delta C = 0.0766$), driven by the age/usage group. Correlated age-proxy features can therefore share a shortcut that leave-one-out misses: the pre-registered single-feature null stands, but the residual age/usage signal remains detectable as a cluster.

\paragraph{What this means.}
The LOO null indicates that no \emph{single} SMART attribute is a near-tautological driver of $C = 0.9595$. In an earlier iteration fit on an unfiltered drive population, power-on hours alone accounted for a 0.28 drop when removed; correcting the cohort eliminated that single-feature concentration. The cluster extension shows that an age/usage \emph{group} still carries a substantial share of discrimination ($\Delta C \approx 0.09$) after the cohort correction -- invisible to LOO, visible when the proxies are removed together. The conditional D-Calibration rejection in Subsection~\ref{subsec:h2} therefore coexists with residual age-proxy concentration at the cluster level, not with a fully distributed feature set. Good discrimination -- even when partly age-mediated across correlated sensors -- still does not imply good calibration.

\FloatBarrier
\subsection{H5 -- Increase in absolute prediction error at later horizons}
\label{subsec:h5}

Among the Domain~3 forests, \texttt{rsf\_content} has global $C=0.6982$ within the pre-registered band and IPCW Brier scores of 0.1633, 0.1851, and 0.1935 at 12, 24, and 36 months. The pre-registered rule rejects (Holm-adjusted $p=0.0040$), establishing higher absolute prediction error at later horizons. Because each horizon defines a different target, the raw increase alone does not establish relative deterioration.

\begin{center}
\includegraphics[width=0.88\linewidth,height=0.50\textheight,keepaspectratio]{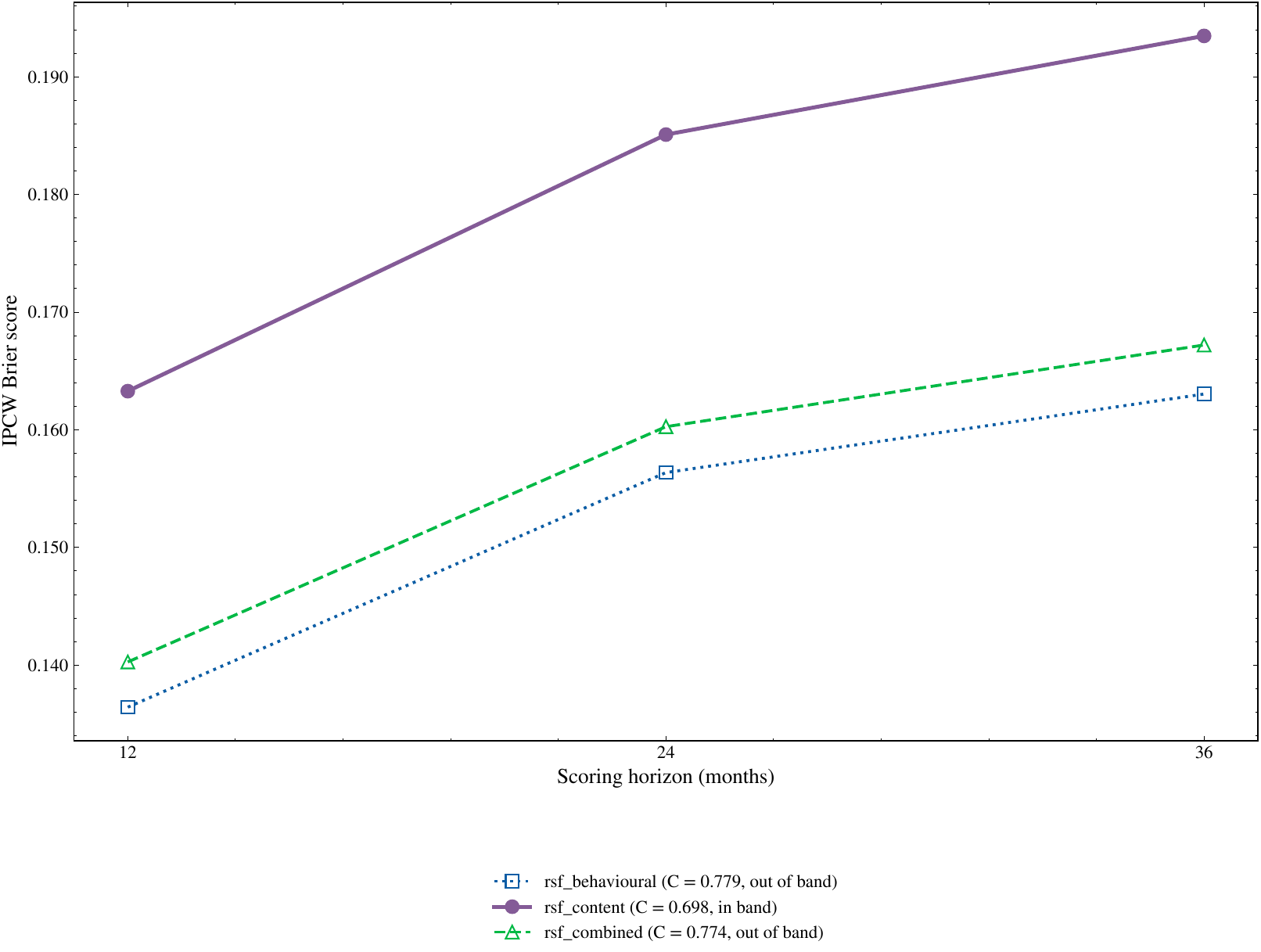}
\captionof{figure}{IPCW Brier score at three horizons. \texttt{rsf\_content} is the only model in the pre-registered C-index band and satisfies H5. The figure shows higher absolute prediction error at later horizons; it does not compare performance with a horizon-specific reference.}
\label{fig:h5}
\end{center}

The post-hoc reference-relative analysis gives model/reference Brier values of 0.1633/0.1888, 0.1851/0.2133, and 0.1935/0.2264, hence BSS values of 0.1353, 0.1322, and 0.1454. The paired-bootstrap contrast BSS(36)$-$BSS(12) is 0.0101 (95\% CI $[-0.0082,0.0276]$). This is scenario~C: absolute Brier error is higher at later horizons, but relative model deterioration is not supported. The originally planned temporal-AUC sensitivity is non-monotonic (0.7295, 0.7171, 0.7272) and does not support temporal decline.

\begin{figure}[!htbp]
\centering
\includegraphics[width=0.95\linewidth,height=0.50\textheight,keepaspectratio]{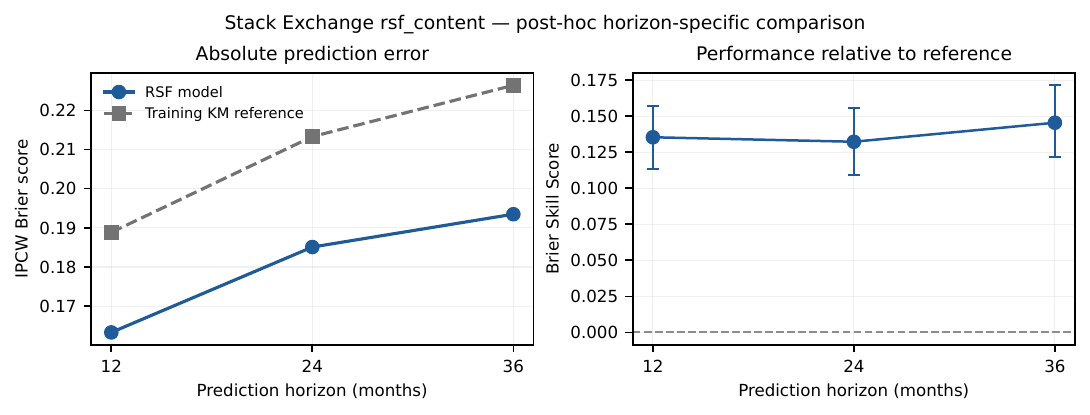}
\caption{Post-hoc H5 reference comparison for \texttt{rsf\_content}. Model and training-set marginal Kaplan--Meier reference Brier scores appear at left; BSS with paired-bootstrap 95\% intervals ($B=1{,}000$) appears at right. Positive skill at all horizons and the 36-minus-12 contrast do not support relative deterioration.}
\label{fig:h5b}
\end{figure}

\FloatBarrier
\subsection{Summary of hypothesis-specific results}
\label{subsec:verdict}

Table~\ref{tab:verdict} reports H1--H5 separately after outer Holm correction. H2, H3, and H5 reject under the literal protocol reading; H1 and H4 do not. A proportional reading of H3 requires five of seven strata and leaves only H2 and H5 rejecting. The resulting 3/5 literal and 2/5 proportional counts are descriptive aggregations only. The frozen count rule does not establish a calibrated omnibus test, so we assign no global $p$-value or inferential verdict.

\begin{table}[!htbp]
\centering
\caption{Holm-adjusted hypothesis-specific results. Literal and proportional counts are descriptive only; no omnibus inference is assigned.}
\label{tab:verdict}
\begin{threeparttable}
\begin{tabular}{lccc}
\toprule
\textbf{Hypothesis} & \textbf{Raw $p$} & \textbf{Holm-adjusted $p$} & \textbf{Reject $H_0$} \\
\midrule
H1 & 0.0779 & 0.1558 & No \\
H2 & $2.60\times10^{-136}$ & $1.30\times10^{-135}$ & Yes \\
H3 & 0.0149 & 0.0448 & Yes (literal) \\
H4 & 1.000 & 1.000 & No \\
H5 & 0.0010 & 0.0040 & Yes \\
\bottomrule
\end{tabular}
\begin{tablenotes}\footnotesize
\item H4 uses 1.000 by convention for its threshold/existence decision. Under the proportional H3 sensitivity, H3 does not reject.
\end{tablenotes}
\end{threeparttable}
\end{table}

The outcomes have distinct qualifications: H2 is conditional and in-sample; H3 is an estimator-level competing-risks audit and depends on literal versus proportional interpretation; H5 establishes an absolute-error increase while its post-hoc BSS does not support relative deterioration. Together they show dimensions not captured by the global C-index in the evaluated models and domains.

\FloatBarrier
\subsection{Sensitivity to censoring dependence (Domain 1)}
\label{subsec:sensitivity}

The standard procedures used for the five pre-registered hypotheses require non-informative censoring for their stated interpretations. The anchor paper's central argument is that this assumption is often violated, and that violation biases evaluation. We validated the anchor's copula-aware tools on synthetic data (Subsection~\ref{subsec:instrument-validation}) but did not include them in the pre-registered battery because the dependence parameter is fundamentally unidentifiable from the data: no test can tell us how strongly censoring depends on failure risk in the Backblaze dataset.

\begin{figure}[!htbp]
\centering
\includegraphics[width=0.95\linewidth,height=0.42\textheight,keepaspectratio]{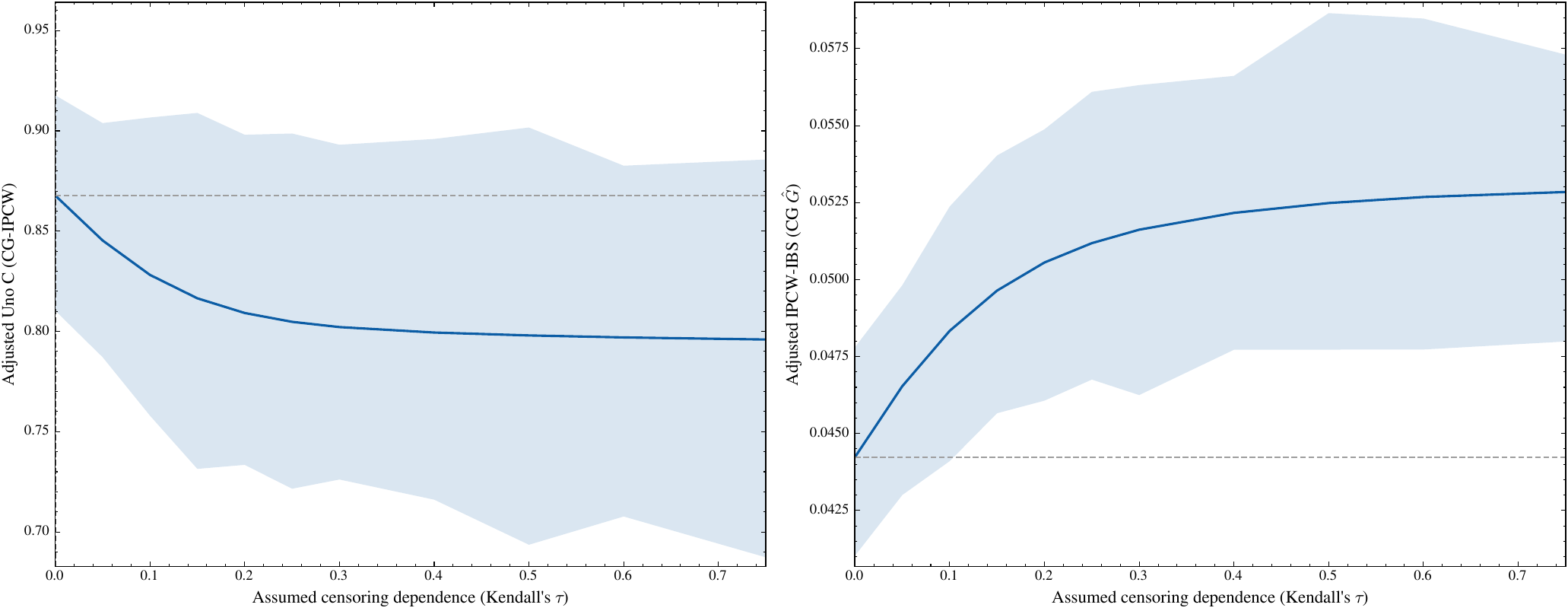}
\caption{Sensitivity of CG-IPCW evaluation metrics to the assumed strength of censoring dependence (Kendall's $\tau$) for the Domain 1 Cox model. Adjusted metrics replace the Kaplan-Meier censoring estimator inside IPCW weights with the Copula-Graphic estimator (Clayton; $\alpha = 2\tau/(1-\tau)$). The vertical dashed line at $\tau = 0$ is independent (standard) censoring; $\tau = 0.75$ is the strongest dependence tested in the anchor's synthetic experiments. Horizontal dashed lines are the unadjusted CG-IPCW references at $\tau = 0$: Uno $C=0.868$, IPCW-IBS $=0.044$. Shaded bands are 95\% bootstrap confidence intervals ($B = 200$, stratified by event). See Subsection~\ref{subsec:sensitivity}.}
\label{fig:copula-sweep}
\end{figure}

What we can do instead is ask how sensitive our conclusions are to this unknown quantity. Figure~\ref{fig:copula-sweep} sweeps Kendall's $\tau$ -- the strength of assumed censoring dependence -- from 0 (the standard independent-censoring assumption) to 0.75 (the strongest dependence the anchor paper tested) and plots Copula-Graphic IPCW-adjusted Uno's C and Integrated Brier Score for the Domain 1 Cox model at each value. Operationally, the Copula-Graphic estimator replaces the Kaplan-Meier estimator of the censoring survival $G(t)$ inside the IPCW weights; the resulting curves are therefore Uno's C and IPCW-IBS under CG-based $\hat{G}$, reducing to the standard KM-IPCW estimators at $\tau = 0$ (verified in-script against the same inputs). The paper's headline Harrell C of 0.9595 does not use IPCW and is not itself swept.

Beyond $\tau \approx 0.25$, adjusted Uno's C declines from 0.868 to about 0.80 and then plateaus through $\tau = 0.75$; adjusted IBS rises modestly from 0.044 to 0.053. The \emph{magnitude} of the IPCW-based metrics is therefore sensitive to the dependence assumption -- underscoring the anchor paper's point that assumption alignment matters -- but even under the strongest dependence we test, adjusted Uno's C remains near 0.80, far from chance-level ranking. Domains 2 and 3 are left for future work.

\FloatBarrier
\section{Discussion}
\label{sec:discussion}

\subsection{Where the C-index is incomplete}
\label{subsec:where-illusion}

Taken separately, the five hypotheses map where C-index-only evaluation is informative and where it is incomplete in this study. H1 does not establish ranking inversion, and H4 finds no single-feature ablation hit, although a post-hoc age/usage cluster does. H2 rejects an in-sample calibration test conditional on non-informative censoring. H3 finds competing-risks bias under the literal frozen rule but not its proportional sensitivity. H5 finds higher absolute prediction error at later horizons, while the post-hoc BSS and originally planned temporal-AUC sensitivity do not support relative deterioration. These findings concern the evaluated models and three domains; they are not an omnibus statement about non-clinical survival ML.

The practical implication is correspondingly specific. A global C-index can coexist with calibration discrepancies, competing-risk bias, or horizon-specific absolute error that it is not designed to measure. Reporting calibration and proper scores alongside discrimination exposes those dimensions, but their interpretation must retain the assumptions and validation status of each procedure.

\subsection{What the reproduction process itself revealed}
\label{subsec:repro-revealed}

Two findings emerged not from the pre-registered hypotheses but from the process of reproducing the baselines faithfully, and both are worth surfacing on their own terms. First, every one of the seven models across all three domains -- not only the one hypothesis formally targets -- fails D-Calibration. We did not pre-register this as a claim, because doing so after seeing the pattern would be exactly the kind of post-hoc inflation this paper argues against; we report it descriptively, as a pattern consistent with, but broader than, what H2 formally establishes.

Second, an earlier iteration of our Domain 1 pipeline -- fit on the full, unfiltered population of drives rather than the cohort the original paper actually used -- reproduced a C-index of 0.981 rather than 0.958, and in that setting, a single feature (power-on hours, a direct proxy for drive age) accounted for a 0.28 drop in discrimination when ablated. The model, in other words, was substantially a restatement of ``older drives fail more often.'' Once we identified and corrected the cohort definition -- healthy drives filtered to a calendar age above seven years, combined with all failed drives regardless of age, which is what the original paper's stated filter actually implies -- that single-feature concentration vanished. A subsequent cluster ablation (Subsection~\ref{subsec:h4}) shows, however, that correlated age/usage proxies (SMART~9, 240, 241) still jointly carry $\Delta C \approx 0.09$ under the same decision rule -- a residual shortcut invisible to leave-one-out. We flag both findings as a methodological caution: conclusions from feature-ablation studies can be highly sensitive to cohort composition and to whether correlated features are removed singly or as a group.

Finally, in validating our evaluation instrument against the anchor's own synthetic experiment (Subsection~\ref{subsec:instrument-validation}), we noted that two open-source implementations of Uno's IPCW-adjusted C-index -- \texttt{SurvivalEVAL} and \texttt{scikit-survival} -- produce measurably different results specifically under dependent censoring, converging only once we aligned our harness to the same backend the anchor's own notebook uses. Under $\tau = 0.50$, SurvivalEVAL reports Uno's C bias of $0.1349$ while \texttt{scikit-survival} reports $0.0656$ ($\Delta = +0.0694$). The mechanism is instructive: Uno's C relies on inverse-probability-of-censoring weights, which in turn require an estimate of the censoring distribution. The two libraries differ in how they estimate that distribution -- how they handle tied event times, how they truncate extreme weights, and how they construct the Kaplan-Meier estimator of censoring. Under independent censoring these differences are negligible; under dependent censoring, where IPCW weights become larger and more variable, they amplify into metric-level disagreement. In practical terms, two researchers evaluating the same model on the same data with different Python imports can arrive at different Uno's C values -- and neither would know, from the metric alone, that the discrepancy is a software artifact rather than a modeling finding. This is a small-scale, practitioner-level echo of the anchor paper's central point: even the choice of software implementing a metric can shift results once the assumption the metric was built for is violated.

\subsection{Limitations}
\label{subsec:limitations}

Several limitations qualify how far these results should be read.

The test of ranking inversion (H1) is under-powered by design in Domain 2, where only two models are compared. A confidence interval on Kendall's $\tau$ computed from two rankings necessarily spans the entire possible range, so the qualitative inversion we observe there cannot be distinguished from chance under our pre-registered decision rule. This is a limitation of the available baseline models, not of the test itself, and it means H1's non-rejection should be read as inconclusive on this domain specifically, rather than as evidence against inversion in general.

Three domains do not constitute a representative survey. We identify reproducible instances of the evaluated discrepancies outside clinical and synthetic settings, but our design cannot estimate how frequently they occur across non-clinical survival ML.

Reproduction is not replication. Domain 1 now reproduces to within 0.0015 of the published C-index, but Domain 2's boosted model remains at an approximate match (gap of 0.0236), isolated to boosting hyperparameters the original paper does not fully specify. Domain 1's reproduced cohort counts (12{,}815 healthy, 5{,}089 failed) differ from the paper's reported figures (12{,}993 and 4{,}889) by approximately 200 drives even after matching the stated filter; this residual ($\sim$1.4\% of the total cohort) is small and unlikely to affect the H4 ablation null result, though we did not formally test this. Where a gap persists, we report it rather than treat it as resolved, and the reproduction protocol (Subsection~\ref{subsec:phase-a}) was itself fixed only after observing that all three domains cleared it -- a decision we disclose rather than obscure, and one that we verified does not affect any hypothesis's outcome, since it governs baseline admission rather than test results.

Excluding the one approximate-match model (D2 boosted Cox, gap 0.0236) from the analysis does not change any hypothesis verdict: H1's Domain 2 inversion is driven by the comparison between \texttt{cox\_classical} and \texttt{cox\_xgboost}, both of which enter the study under the strict tier; H3 and H5 do not involve Domain 2's boosted model.

Our validation of the evaluation instrument (Subsection~\ref{subsec:instrument-validation}) establishes fidelity of integration with the anchor's own code, not an independently verified re-implementation of its metrics. A latent error in the anchor's implementation, if one exists, would be inherited by our harness rather than caught by this validation.

H3's rejection is narrow in two respects. The observed bias (0.0218) exceeds the pre-registered threshold (0.02) by only 0.0018. More consequentially, the freeze specifies ``$\geq 3$ of 5 rating strata'' -- a count written without first inspecting the data, which contain seven ratings rather than five. We applied the frozen absolute count ($\geq 3$); a fractional reading ($\geq 60\%$, requiring $\geq 5$ of 7) is equally defensible and would not reject H3. Under that reading, only H2 and H5 reject, yielding the descriptive count 2/5. The correct practice would have been to verify the number of strata before freezing the protocol; we failed to do so and report both readings rather than resolving the ambiguity in our favor. The underlying estimator finding -- directional bias growing monotonically with credit risk -- does not depend on which reading is adopted. Its scope is deliberately procedural: it concerns the cumulative-incidence estimator used to interpret default risk under competing events, not the Cox models' individual predictions.

D-Calibration and the post-hoc fixed-horizon procedures require non-informative censoring. Under dependent censoring, some apparent calibration discrepancy could reflect metric--assumption misalignment. The size of the D-Calibration $p$-value does not validate the test when this assumption fails. Both analyses use the model-fit cohort and are goodness-of-fit assessments rather than out-of-sample validation. The OOS gate produced no estimate because the train-only Cox refit did not converge under the frozen model choices.

More broadly, we applied the anchor's Copula-Graphic IPCW metrics to Domain 1 as a sensitivity sweep over the unidentifiable dependence parameter (Subsection~\ref{subsec:sensitivity}); Domains 2 and 3 remain as future work. We validated the underlying CG implementation against the anchor's synthetic experiment (Subsection~\ref{subsec:instrument-validation}) before deploying that arm of the ladder on real data.

\subsection{Future work}
\label{subsec:future-work}

The most direct extension of this study is scale: more domains and, within each domain, more competing published models, would give the ranking-inversion test the statistical power it currently lacks in at least one domain. Extending the Domain~1 Copula-Graphic IPCW sensitivity sweep of Subsection~\ref{subsec:sensitivity} to Domains 2 and 3 would close the remaining real-data gap for the dependent-censoring arm of the ladder. Extending the audit to deep survival models (DeepSurv, DeepHit) rather than Cox and random survival forests would test whether the dissociation between discrimination and calibration we observe is specific to these model classes or more general. An independent re-implementation of the calibration and proper-scoring metrics -- rather than the ported, integration-validated implementation used here -- would close the one gap our own validation cannot: verifying the anchor's code against a second, independently written instrument. We release the harness built for this study specifically so that these extensions do not require rebuilding the pipeline from scratch.

\section{Conclusion}
\label{sec:conclusion}

The ICML 2026 Spotlight paper \emph{"Stop Chasing the C-index"} argued on synthetic data that discrimination alone can provide an incomplete assessment of survival models. Across three published non-clinical models, our hypothesis-specific tests identify several such omissions. H1 and H4 do not reject. H2 rejects calibration for Domain~1 conditional on non-informative censoring, but remains an in-sample goodness-of-fit result. H3 identifies competing-risks estimator bias under the literal frozen rule, while a proportional reading does not reject. H5 establishes higher absolute prediction error at later horizons; its post-hoc BSS comparison and the originally planned temporal-AUC sensitivity do not support relative deterioration.

The evidence therefore supports a narrower conclusion than a global rejection claim: in the evaluated models and domains, the global C-index did not capture every decision-relevant dimension examined. Calibration tests, competing-risk analyses, and horizon-specific proper scores provide complementary information, subject to their assumptions. More domains and genuinely out-of-sample calibration analyses are needed before generalizing frequency or magnitude beyond these cases.

\section*{Reproducibility statement}

All code, data-loading scripts, and the exact reproduction and evaluation pipeline used in this paper are released at \url{https://github.com/rafa-rodriguess/c_index_illusion_pub} under an open license. The repository is organized as a single orchestrator notebook, \texttt{00\_pipeline\_notebook.ipynb}, which runs the full study end to end -- baseline reproduction for all three domains, anchor-instrument validation, the evaluation ladder, all five hypothesis tests, and their Holm-adjusted hypothesis-specific summary -- with inline commentary at each stage explaining what is computed and why. We adopted this single-notebook structure deliberately, over a more conventional multi-script package, so that a reviewer or future user can step through the entire pipeline in the order it was actually executed, rather than reconstructing that order from a collection of separate files.

The protocol freeze (see Appendix~\ref{app:protocol}) documents every methodological decision made before any test data were examined, including the two decisions we revised after initial results and disclose as such rather than silently correcting: the definition of Domain 1's cohort filter (Subsection~\ref{subsec:results-repro}, Subsection~\ref{subsec:repro-revealed}) and the numerical reproduction tolerance (Subsection~\ref{subsec:phase-a}), neither of which affects any hypothesis's outcome. All bootstrap procedures use a fixed random seed (seed $= 42$, $B = 1{,}000$ resamples throughout) and are re-runnable from the released code. All three underlying datasets -- Backblaze Drive Stats, the Bondora peer-to-peer loan book, and the Stack Exchange data dump -- are public and their exact access points are documented in Appendix~\ref{app:reproduction} alongside the per-domain preprocessing decisions required to reproduce each baseline.

We followed the reproducibility checklist proposed by \citet{pineau2021} in preparing this release.

\section*{Disclosure of generative AI use}

Generative AI tools were used in the preparation of this paper in two capacities, both disclosed here for transparency. Claude (Anthropic) was used to refine prose and to translate portions of the manuscript, originally drafted in Portuguese, into English; all scientific claims, hypotheses, analyses, and conclusions originate from the author and were verified against the pipeline's output before inclusion. Consensus (consensus.app), an AI-assisted academic search engine, was used to help locate and triage candidate literature during the related-work search; every citation retrieved this way was independently read and verified against the primary source before being cited. No generative AI tool was used to produce, analyze, or interpret any of the results reported in this paper.

\subsubsection*{Acknowledgments}
This paper exists because of work we did not do ourselves. We thank Christian Marius Lillelund, Shi-ang Qi, Russell Greiner, and Christian Fischer Pedersen for the position paper this study responds to, and for releasing the code and synthetic experiment that made independent validation of our own instrument possible -- the kind of transparency that empirical follow-up work like ours depends on and too rarely receives. We thank Jishan Ahmed and Robert C.\ Green II, Gerrit Ferdinand Bone-Winkel and Falko Reichenbach, and Hooman Abedi Firouzjaei for the three baseline studies audited here; reproducing their work closely enough to subject it to further scrutiny is only possible because they described it clearly enough, and released their data openly enough, to be reproduced at all. Science of the kind this paper attempts -- testing a claim by building directly on top of it -- is only possible when the work underneath is generous with its methods. We hope this paper is read in that spirit, and extended by others in turn. Funding: none declared. Competing interests: none declared.

\bibliography{refs}

@Article{emmertstreib2019,
  author  = {Emmert-Streib, Frank and Dehmer, Matthias},
  title   = {Introduction to Survival Analysis in Practice},
  journal = {Machine Learning and Knowledge Extraction},
  year    = {2019},
  volume  = {1},
  number  = {3},
  pages   = {1013--1038},
  doi     = {10.3390/make1030058},
  url     = {https://doi.org/10.3390/make1030058}
}

@Article{wang2019,
  author  = {Wang, Ping and Li, Yan and Reddy, Chandan K.},
  title   = {Machine Learning for Survival Analysis: A Survey},
  journal = {ACM Computing Surveys},
  year    = {2019},
  volume  = {51},
  number  = {6},
  pages   = {110:1--110:36},
  doi     = {10.1145/3214306},
  url     = {https://doi.org/10.1145/3214306}
}

@misc{lillelund2025b,
  author       = {Lillelund, Christian Marius and Qi, Shi-ang and Greiner, Russell and Pedersen, Christian Fischer},
  title        = {Position: Stop Chasing the {C}-index when Evaluating Survival Analysis Models},
  year         = {2025},
  eprint       = {2506.02075},
  archivePrefix= {arXiv},
  primaryClass = {cs.LG},
  url          = {https://arxiv.org/abs/2506.02075}
}

@misc{lillelund2025a,
  author       = {Lillelund, Christian Marius and Qi, Shi-ang and Greiner, Russell},
  title        = {Overcoming Dependent Censoring in the Evaluation of Survival Models},
  year         = {2025},
  eprint       = {2502.19460},
  archivePrefix= {arXiv},
  primaryClass = {stat.ML},
  url          = {https://arxiv.org/abs/2502.19460}
}

@Article{wissel2025,
  author  = {Wissel, David and Janakarajan, Nikita and Grover, Aayush and Toniato, Elisa and Rodr{\'i}guez Mart{\'i}nez, Mar{\'i}a and Boeva, Valentina},
  title   = {{SurvBoard}: Standardized Benchmarking for Multi-omics Cancer Survival Models},
  journal = {Briefings in Bioinformatics},
  year    = {2025},
  volume  = {26},
  number  = {5},
  pages   = {bbaf521},
  doi     = {10.1093/bib/bbaf521},
  url     = {https://doi.org/10.1093/bib/bbaf521}
}

@Article{harrell1982,
  author  = {Harrell, Frank E., Jr. and Califf, Robert M. and Pryor, David B. and Lee, Kerry L. and Rosati, Robert A.},
  title   = {Evaluating the Yield of Medical Tests},
  journal = {{JAMA}},
  year    = {1982},
  volume  = {247},
  number  = {18},
  pages   = {2543--2546},
  doi     = {10.1001/jama.1982.03320430047030},
  url     = {https://doi.org/10.1001/jama.1982.03320430047030}
}

@Article{uno2011,
  author  = {Uno, Hajime and Cai, Tianxi and Pencina, Michael J. and D'Agostino, Ralph B. and Wei, L. J.},
  title   = {On the {C}-statistics for Evaluating Overall Adequacy of Risk Prediction Procedures with Censored Survival Data},
  journal = {Statistics in Medicine},
  year    = {2011},
  volume  = {30},
  number  = {10},
  pages   = {1105--1117},
  doi     = {10.1002/sim.4154},
  url     = {https://doi.org/10.1002/sim.4154}
}

@Article{antolini2005,
  author  = {Antolini, Laura and Boracchi, Patrizia and Biganzoli, Elia},
  title   = {A Time-dependent Discrimination Index for Survival Data},
  journal = {Statistics in Medicine},
  year    = {2005},
  volume  = {24},
  number  = {24},
  pages   = {3927--3944},
  doi     = {10.1002/sim.2427},
  url     = {https://doi.org/10.1002/sim.2427}
}

@Article{rossi2025,
  author  = {Rossi, Irene and Sartori, Federico and Rollo, Corrado and Birolo, Giovanni and Fariselli, Piero and Sanavia, Tiziana},
  title   = {Beyond {Cox} Models: Assessing the Performance of Machine-learning Methods in Non-proportional Hazards and Non-linear Survival Analysis},
  journal = {Computers in Biology and Medicine},
  year    = {2025},
  volume  = {198},
  pages   = {111176},
  doi     = {10.1016/j.compbiomed.2025.111176},
  url     = {https://doi.org/10.1016/j.compbiomed.2025.111176}
}

@Article{haider2020,
  author  = {Haider, Humza and Hoehn, Bret and Davis, Sarah and Greiner, Russell},
  title   = {Effective Ways to Build and Evaluate Individual Survival Distributions},
  journal = {Journal of Machine Learning Research},
  year    = {2020},
  volume  = {21},
  number  = {85},
  pages   = {1--63},
  url     = {http://jmlr.org/papers/v21/18-772.html}
}

@Article{graf1999,
  author  = {Graf, Erika and Schmoor, Claudia and Sauerbrei, Willi and Schumacher, Martin},
  title   = {Assessment and Comparison of Prognostic Classification Schemes for Survival Data},
  journal = {Statistics in Medicine},
  year    = {1999},
  volume  = {18},
  number  = {17-18},
  pages   = {2529--2545},
  doi     = {10.1002/(SICI)1097-0258(19990915/30)18:17/18<2529::AID-SIM274>3.0.CO;2-5},
  url     = {https://doi.org/10.1002/(SICI)1097-0258(19990915/30)18:17/18<2529::AID-SIM274>3.0.CO;2-5}
}

@Article{coemans2022,
  author  = {Coemans, Maarten and Verbeke, Geert and D{\"o}hler, Bernd and S{\"u}sal, Caner and Naesens, Maarten},
  title   = {Bias by Censoring for Competing Events in Survival Analysis},
  journal = {{BMJ}},
  year    = {2022},
  volume  = {378},
  pages   = {e071349},
  doi     = {10.1136/bmj-2022-071349},
  url     = {https://doi.org/10.1136/bmj-2022-071349}
}

@Article{satagopan2004,
  author  = {Satagopan, Jaya M. and Ben-Porat, Leah and Berwick, Marianne and Robson, Mark and Kutler, David and Auerbach, Arleen D.},
  title   = {A Note on Competing Risks in Survival Data Analysis},
  journal = {British Journal of Cancer},
  year    = {2004},
  volume  = {91},
  number  = {7},
  pages   = {1229--1235},
  doi     = {10.1038/sj.bjc.6602102},
  url     = {https://doi.org/10.1038/sj.bjc.6602102}
}

@Article{fine1999,
  author  = {Fine, Jason P. and Gray, Robert J.},
  title   = {A Proportional Hazards Model for the Subdistribution of a Competing Risk},
  journal = {Journal of the American Statistical Association},
  year    = {1999},
  volume  = {94},
  number  = {446},
  pages   = {496--509},
  doi     = {10.1080/01621459.1999.10474144},
  url     = {https://doi.org/10.1080/01621459.1999.10474144}
}

@Article{vangeloven2022,
  author  = {van Geloven, Nan and Giardiello, Daniele and Bonneville, Edouard F. and Teece, Lucy and Ramspek, Chava L. and van Smeden, Maarten and Snell, Kym I. E. and van Calster, Ben and Pohar-Perme, Maja and Riley, Richard D. and Putter, Hein and Steyerberg, Ewout W.},
  title   = {Validation of Prediction Models in the Presence of Competing Risks: A Guide Through Modern Methods},
  journal = {{BMJ}},
  year    = {2022},
  volume  = {377},
  pages   = {e069249},
  doi     = {10.1136/bmj-2021-069249},
  url     = {https://doi.org/10.1136/bmj-2021-069249}
}

@misc{jeanselme2025,
  author       = {Jeanselme, Vincent and Tom, Brian and Barrett, Jessica},
  title        = {Competing Risks: Impact on Risk Estimation and Algorithmic Fairness},
  year         = {2025},
  eprint       = {2508.05435},
  archivePrefix= {arXiv},
  primaryClass = {stat.ML},
  url          = {https://arxiv.org/abs/2508.05435}
}

@Article{frydman2020,
  author  = {Frydman, Halina and Matuszyk, Anna},
  title   = {Random Survival Forest for Competing Credit Risks},
  journal = {Journal of the Operational Research Society},
  year    = {2020},
  doi     = {10.1080/01605682.2020.1759385},
  url     = {https://doi.org/10.1080/01605682.2020.1759385}
}

@Article{wycinka2019,
  author  = {Wycinka, Ewa},
  title   = {Competing Risk Models of Default in the Presence of Early Repayments},
  journal = {Econometrics. Ekonometria. Advances in Applied Data Analysis},
  year    = {2019},
  volume  = {23},
  number  = {2},
  pages   = {99--118},
  doi     = {10.15611/eada.2019.2.07},
  url     = {https://doi.org/10.15611/eada.2019.2.07}
}

@Article{ahmed2024,
  author  = {Ahmed, Junaid and Green, Roger},
  title   = {Leveraging Survival Analysis in Cost-aware {DeepNet} for Efficient Hard Drive Failure Prediction},
  journal = {Neural Computing and Applications},
  year    = {2024},
  doi     = {10.1007/s00521-024-10479-6},
  url     = {https://doi.org/10.1007/s00521-024-10479-6}
}

@Article{bonewinkel2024,
  author  = {Bone-Winkel, Georg F. and Reichenbach, Felix},
  title   = {Improving Credit Risk Assessment in {P2P} Lending with Explainable Machine Learning Survival Analysis},
  journal = {Digital Finance},
  year    = {2024},
  doi     = {10.1007/s42521-024-00114-3},
  url     = {https://doi.org/10.1007/s42521-024-00114-3}
}

@Article{abedi2022,
  author  = {Abedi Firouzjaei, Hassan},
  title   = {Survival Analysis for User Disengagement Prediction: Question-and-answering Communities' Case},
  journal = {Social Network Analysis and Mining},
  year    = {2022},
  volume  = {12},
  pages   = {45},
  doi     = {10.1007/s13278-022-00914-8},
  url     = {https://doi.org/10.1007/s13278-022-00914-8}
}

@Article{pineau2021,
  author        = {Pineau, Joelle and Vincent-Lamarre, Philippe and Sinha, Koustuv and Larivi{\`e}re, Vincent and Beygelzimer, Alina and d'Alch{\'e}-Buc, Florence and Fox, Emily and Larochelle, Hugo},
  title         = {Improving Reproducibility in Machine Learning Research (A Report from the {NeurIPS} 2019 Reproducibility Program)},
  journal       = {Journal of Machine Learning Research},
  year          = {2021},
  volume        = {22},
  number        = {164},
  pages         = {1--20},
  eprint        = {2003.12206},
  archivePrefix = {arXiv},
  primaryClass  = {cs.LG},
  url           = {https://arxiv.org/abs/2003.12206}
}
\bibliographystyle{tmlr}

\appendix

\section{Protocol Freeze}
\label{app:protocol}

Protocol Freeze Record (version \texttt{2026-07-12.c00.v5.1}). Every methodological decision documented here was fixed before any test data were examined, except where explicitly noted (C00.4, Subsection~\ref{subsec:phase-a}). Rejected alternatives are listed for transparency.

\begin{itemize}
\item Frozen at (UTC): \texttt{2026-07-12T04:05:29.991888+00:00}
\item Content SHA-256: \texttt{2fae30d94c79f5b8e245ea40a884bf08}\\
\texttt{cd1758ffe768f86891e05f781de96cb0}
\item Random seed: \texttt{42}
\item $\alpha = 0.05$, $B = 1000$, permutations $= 10000$
\end{itemize}

\subsection*{C00.1 -- H1 decision rule}
Reject $H_0$ in a domain iff observed $\tau_K \leq 0.5$ AND the subject-level stratified bootstrap ($B{=}1000$) $p$-value for $H_0{:}\ \tau_K = 1$ is $< \alpha$, with $p = (1 + \#\{\tau_b \geq 1\}) / (B + 1)$. Percentile CI for $\tau_K$ is reported; rank-permutation is sensitivity-only (unreachable for $k\leq 3$).
Binary inversion cross-domain: \textbf{descriptive\_only}.

\subsection*{C00.2 -- H1 ranking objects}
\begin{center}
\begin{tabular}{ll}
\toprule
\textbf{Domain} & \textbf{Models} \\
\midrule
domain1 & \texttt{cox\_full}, \texttt{cox\_ablated} \\
domain2 & \texttt{cox\_classical}, \texttt{cox\_xgboost} \\
domain3 & \texttt{rsf\_behavioural}, \texttt{rsf\_content}, \texttt{rsf\_combined} \\
\bottomrule
\end{tabular}
\end{center}

\subsection*{C00.3 -- Multiple-testing family}
Inner: Holm within hypothesis to collapse to one decision $p$ (H1: domains; H3: rating strata; H2/H4/H5: already single). Outer: Holm--Bonferroni over the 5 primary hypothesis $p$-values. Meta family size: 5.

\emph{Post-review interpretation.} This historical rule is preserved verbatim below, but the revised manuscript does not treat H$_{\mathrm{meta}}$ as a calibrated omnibus test. H1--H5 are interpreted separately; 3/5 under the literal H3 reading and 2/5 under the proportional reading are descriptive aggregations without a global $p$-value or inferential verdict. The frozen H5 label is likewise historical; the revised text interprets it as an increase in absolute prediction error, not relative model deterioration.

\subsection*{Hypotheses (decision rules)}

\paragraph{H1 -- Ranking inversion under metric-assumption misalignment.}
\textbf{$H_0$:} Within each domain, C-index ranking equals IPCW-IBS ranking ($\tau_K = 1$).
\textbf{$H_1$:} $\tau_K < 1$ in at least one domain (math alternative; decision = C00.1).
\textbf{Statistic:} Kendall $\tau_K$ (C-index rank vs.\ IPCW-IBS rank); subject-level stratified bootstrap CI (primary); rank permutation (sensitivity).
\textbf{Decision:} C00.1.

\paragraph{H2 -- Discrimination-reliability dissociation.}
\textbf{$H_0$:} Every model with $C_H \geq 0.90$ passes D-Calibration ($p \geq 0.05$).
\textbf{$H_1$:} Exists model with $C_H \geq 0.90$ and $p_{\mathrm{D\text{-}cal}} < 0.05$.
\textbf{Statistic:} D-Calibration $\chi^2$, 10 quantile bins \citep{haider2020}.
\textbf{Decision:} Reject if Backblaze reproduction with $C_H\geq 0.90$ has $p_{\mathrm{D\text{-}cal}} < 0.05$.

\paragraph{H3 -- Directional bias from competing-risks blindness.}
\textbf{$H_0$:} $|F_{\mathrm{naive,default}}(12\mathrm{m}) - F_{\mathrm{AJ,default}}(12\mathrm{m})| \leq 0.02$ and bootstrap CI contains 0.
\textbf{$H_1$:} Absolute difference $> 0.02$, CI excludes 0; predicted direction $F_{\mathrm{naive}} > F_{\mathrm{AJ}}$.
\textbf{Statistic:} Absolute CIF difference at 12 months; $B{=}1000$ percentile CI.
\textbf{Decision:} Reject if $\Delta > 0.02$ AND positive sign in $\geq 3$ of 5 rating strata.

\emph{Disclosure (strata count).} The freeze text and PROTOCOL field \texttt{n\_rating\_strata}${}={}$5 anticipated five Bondora ratings. The baseline reports seven (AA--F), and all seven appear in our evaluation data. The operative threshold applied is the absolute count \texttt{min\_strata\_consistent}${}={}$3 (not a $3/5$ fraction of the observed seven). Three of seven strata (D, E, F) met that absolute threshold. We retain the freeze wording and disclose the denominator mismatch here rather than rewriting C00 after seeing the data.

\paragraph{H4 -- Discrimination inflated by a small SMART subset (broad ablation).}
\textbf{$H_0$:} No SMART ablation set yields $\Delta C_H \geq 0.03$ with non-overlapping paired bootstrap CIs (full vs.\ ablated).
\textbf{$H_1$:} There exists a SMART ablation set with $\Delta C_H \geq 0.03$ and non-overlapping CIs -- i.e.\ discrimination is concentrated in a small feature subset (structural leakage / near-failure signal).
\textbf{Statistic:} Broad ablation survey (leave-one-out + reported subsets); paired bootstrap ($B{=}1000$) on $\Delta C_H$ for sets that clear the floor.
\textbf{Decision:} Reject $H_0$ if at least one ablation set meets $\Delta \geq 0.03$ with non-overlapping full vs.\ ablated CIs.

\paragraph{H5 -- Horizon-specific proper-score degradation masked by global C-index.}
\textbf{$H_0$:} IPCW Brier score is non-increasing (or flat) across $t \in \{12,24,36\}$ months, or any rise is within 2 combined SEs.
\textbf{$H_1$:} Brier($t$) increases monotonically on $\{12,24,36\}$; Brier(36)$-$Brier(12) exceeds 2 combined bootstrap SEs; while global Harrell C remains in the reported band $[0.66, 0.76]$.
\textbf{Statistic:} Pointwise IPCW Brier at month horizons (\texttt{sksurv}) + bootstrap SE; time-dependent AUC retained as sensitivity appendix.
\textbf{Decision:} Reject if (for at least one frozen D3 RSF) Brier mono-increases, $\Delta$ Brier(36-12) $> 2$ SE, AND global C in $[0.66, 0.76]$.

\paragraph{H$_{\mathrm{meta}}$ -- C-index alone is not a reliable proxy outside healthcare.}
\textbf{$H_0$:} C-index-only evaluation is a reliable proxy for assumption-aligned evaluation cross-domain outside healthcare.
\textbf{$H_1$:} Rejected if $\geq 3$ of 5 primary hypotheses (H1--H5) are rejected under the outer Holm family of 5 (C00.3).

\subsection*{Reproduction targets}
\begin{center}
{\footnotesize
\begin{tabular}{lp{10cm}}
\toprule
\textbf{Domain} & \textbf{Target} \\
\midrule
domain1 & \texttt{\{"metric": "harrell\_cindex", "reported": 0.958\}} \\
domain2 & \texttt{\{"metric": "rating\_stratification", "reported": "cox\_classical + cox\_xgboost vs Bondora ratings"\}} \\
domain3 & \texttt{\{"metric": "rsf\_cindex\_band", "protocol": "5-fold x 30 runs", "reported\_max": 0.76, "reported\_min": 0.66\}} \\
\bottomrule
\end{tabular}
}
\end{center}

H1 D1 companion ablation SMART ids (\texttt{cox\_ablated} only): $[5, 197, 198]$. Not an H4 claim -- H4 exhibits are LOO survey + H04 bootstrap.

This freeze is authoritative for Blocks D--G. Do not alter it without bumping the protocol version identifier.

\subsection*{C00.4 -- Reproduction admission criterion (declared a posteriori)}
A reproduced quantity within $0.01$ of its reported value is a \emph{strict} match; within $(0.01,0.03]$ is \emph{approximate}; beyond $0.03$ is a \emph{reproduction gap}, reported and retained. This criterion was fixed only after observing that all three domains cleared it, and governs baseline \emph{admission}, not any hypothesis outcome (see Subsection~\ref{subsec:phase-a}).

\emph{Rejected alternative:} pre-registering a numerical tolerance before seeing reproduction gaps -- rejected because none of the baselines pre-specify one; disclosing the a-posteriori choice is required for honesty.

\subsection*{Rejected alternatives (transparency)}
\begin{itemize}
\item \textbf{C00.1:} Rank-permutation as the formal H1 gate -- rejected as unreachable for $k\leq 3$ models; retained as sensitivity-only. Primary test is subject-level stratified bootstrap.
\item \textbf{C00.2:} (B) declare H1 N/A in Domain 1; (C) rank Harrell vs.\ Uno instead of C-index vs.\ IPCW-IBS -- rejected in favour of the three-domain ranking objects above.
\item \textbf{H5:} Uno AUC($t$) decay as primary -- rejected; pointwise IPCW Brier at $\{12, 24, 36\}$ months is primary, with time-dependent AUC as sensitivity.
\item \textbf{H3 strata denominator:} Rewriting the freeze from ``$\geq 3$ of 5'' to ``$\geq 3$ of 7'' after observing seven Bondora ratings -- rejected; we disclose the mismatch and apply the absolute \texttt{min\_strata\_consistent}${}={}$3 threshold instead (see H3 disclosure above).
\end{itemize}

\section{Per-Domain Reproduction Detail}
\label{app:reproduction}

Per-domain reproduction detail: quantities, limitations, and guiding summaries. Referenced from Subsection~\ref{subsec:phase-a}, Subsection~\ref{subsec:results-repro}, and Subsection~\ref{subsec:limitations}.

\subsection{Domain 1 -- Backblaze / Ahmed \& Green (2024)}

\begin{table}[!htbp]
\centering
\caption{All reproduced quantities for Domain 1 -- Backblaze / Ahmed \& Green (2024).}
\label{tab:d01repro}
{\footnotesize
\resizebox{\linewidth}{!}{%
\begin{threeparttable}
\begin{tabular}{lp{2.2cm}p{3.2cm}cc}
\toprule
\textbf{Quantity} & \textbf{Reported} & \textbf{Source} & \textbf{Ours} & \textbf{Gap} \\
\midrule
$N$ drives ST4000DM000 & 37{,}037 & paper \S4 & 37{,}038 & $+1$ \\
$N$ healthy, age $>7$y & 12{,}993 & paper \S4.1 & 12{,}815 & $-178$ \\
$N$ failed, age $>7$y & 4{,}889 & paper \S4.1 & 5{,}089 & $+200$ \\
Harrell C-index & 0.9580 & paper \S7.1 / abstract & 0.9595 & $+0.0015$ \\
HR SMART 184 & 1.01000 & paper \S7.1 & 1.00768 & $-0.00232$ \\
HR SMART 190 & 0.98400 & paper \S7.1 & 0.97282 & $-0.01118$ \\
HR SMART 194 & 0.99000 & paper \S7.1 & n.a. & n.a. \\
$N$ fit complete cases & n.r. & This work & 17{,}826 & n.a. \\
\bottomrule
\end{tabular}
\begin{tablenotes}
    \footnotesize
    \item \textit{Note.} Even after matching the published population filter (ST4000DM000, 2013--2022, 21 SMART raw features) and recovering a Cox C-index within 0.0015 of the headline 0.958, several preprocessing choices required by the baseline remain under-specified and unverifiable because the linked GitLab repository is private. The residual gap is reported as a reproduction finding, not discarded as implementation error.
    \end{tablenotes}
\end{threeparttable}
}
}
\end{table}

\begin{table}[!htbp]
\centering
\caption{Domain 1 limitations: under-specified choices and our resolutions.}
\label{tab:d01lim}
{\footnotesize
\begin{tabular}{@{}c>{\raggedright\arraybackslash}p{2.1cm}>{\raggedright\arraybackslash}p{2.7cm}>{\raggedright\arraybackslash}p{2.7cm}>{\raggedright\arraybackslash}p{2.5cm}@{}}
\toprule
\textbf{\#} & \textbf{Limitation} & \textbf{Paper says / omits} & \textbf{What we did} & \textbf{Implication} \\
\midrule
L1 & Author code inaccessible & Footnote: private GitLab repo (Ahmed \& Green) & Repo private & Best-effort from published text \\
L2 & Cohort ``$>7$ years'' (\S4.1) & 12{,}993 healthy + 4{,}889 failed & Healthy calendar span $>7$y $\cup$ all failed $\rightarrow$ $\sim$12{,}815 / 5{,}089 & Counts $\pm\sim$200; C matches \\
L3 & SMART snapshot for Cox & Silent on first/last/mean & SMART raw on last observed day & May shift C/HRs vs private pipeline \\
L4 & SMART 190 $\equiv$ 194 & HR reported for both & Identical in this fleet $\rightarrow$ drop 194 & Engineering deviation \\
L5 & L2 penalizer & Names CoxPHFitter; silent on regularization & \texttt{penalizer}$=$0.01 & C/HRs depend mildly on this knob \\
L6 & Cox hold-out & S6 80/20 is for DeepNet; S7.1 C as GOF & In-sample C on the fit & Comparable to paper's GOF framing \\
L7 & 15-day horizon & S4.1/S6 labeling for DL classification & Not applied to Cox & Correct for the Cox target \\
\bottomrule
\end{tabular}
}
\end{table}

\FloatBarrier
\subsection{Domain 2 -- Bondora / Bone-Winkel \& Reichenbach (2024)}

\begin{table}[!htbp]
\centering
\caption{All reproduced quantities for Domain 2 -- Bondora / Bone-Winkel \& Reichenbach (2024).}
\label{tab:d02reproapp}
{\footnotesize
\resizebox{\linewidth}{!}{%
\begin{threeparttable}
\begin{tabular}{lp{2.4cm}p{3.0cm}cc}
\toprule
\textbf{Quantity} & \textbf{Reported} & \textbf{Source} & \textbf{Ours} & \textbf{Gap} \\
\midrule
Train/test split date & 2020-01-01 & Baseline \S3.3 & 2020-01-01 & n.a. \\
Test-set size & n.r. & Baseline \S3.3 & 6{,}457 & n.a. \\
Rating strata & 7 & Baseline Table 1 / \S3.5 & 7 & $+0.0000$ \\
Harrell C (linear Cox, test) & 0.6590 & Baseline \S4.1 & 0.6559 & $-0.0031$ \\
Harrell C (boosted Cox, test) & 0.6740 & Baseline \S4.1 & 0.6504 & $-0.0236$ \\
AA default rate, Bondora (completed) & 0.1726 & Baseline Table 1 / footnote 14 & 0.2023 & $+0.0297$ \\
AA default rate, linear Cox (completed) & n.r. & This work & 0.1723 & n.a. \\
AA default rate, boosted Cox (completed) & 0.1445 & Baseline Table 1 / footnote 14 & 0.1486 & $+0.0041$ \\
$N$ loans in boosted AA bucket & 713 & Baseline Table 1 & 467 & $-246$ \\
$N$ loans in linear AA bucket & 858 & Baseline Appendix D & 791 & $-67$ \\
AA default rate, Bondora (KM at term) & 0.1726 & Baseline \S3.6 KM & 0.1706 & $-0.0020$ \\
AA default rate, Bondora (empirical) & 0.1726 & Baseline Table 1 & 0.2102 & $+0.0376$ \\
AA default rate, linear Cox (KM at term) & n.r. & This work & 0.1593 & n.a. \\
AA default rate, boosted Cox (KM at term) & 0.1445 & Baseline Table 1 / \S3.6 KM & 0.1263 & $-0.0182$ \\
IRR, Bondora AA & $-0.0320$ & Baseline Table 1 & n.a. & n.a. \\
IRR, boosted AA & 0.1563 & Baseline Table 1 & n.a. & n.a. \\
\bottomrule
\end{tabular}
\begin{tablenotes}
    \footnotesize
    \item \textit{Note.} n.r.\ = not reported in the baseline; n.a.\ = not applicable (no baseline value to compare, or not computed here). IRR (internal rate of return) requires repayment microdata unavailable in the public dump; linear-Cox AA rates are not published in the baseline main text and are reported from this reproduction only.
    \end{tablenotes}
\end{threeparttable}
}
}
\end{table}

\begin{table}[!htbp]
\centering
\caption{Domain 2 limitations: under-specified choices and our resolutions.}
\label{tab:d02lim}
{\footnotesize
\begin{threeparttable}
\begin{tabular}{@{}c>{\raggedright\arraybackslash}p{2.1cm}>{\raggedright\arraybackslash}p{2.7cm}>{\raggedright\arraybackslash}p{2.7cm}>{\raggedright\arraybackslash}p{2.5cm}@{}}
\toprule
\textbf{\#} & \textbf{Limitation} & \textbf{Paper says / omits} & \textbf{What we did} & \textbf{Implication} \\
\midrule
L1 & No code repository & Zenodo $=$ lifelines & Best-effort from published text & Not bit-exact \\
L2 & Public dump $\neq$ private extract & Retrieved 2024-01-03 & Kaggle + D00 align (as-of 2024-01-03; 36m; 2014--2020) & Schema $\sim$97 vs $\sim$112 cols \\
L3 & 10-step preprocess & S3.2 detailed & \path{domain2_preprocess.py} & Encoding may differ \\
L4 & Repayments / IRR & Table 1 IRR & skipped & Does not block H3 / Phase A \\
L5 & XGB HPO & Optuna + GPU & Optuna TPE 40 trials (CPU hist) & AA boosted $\approx$ paper \\
L6 & Investor xlsx & -- & Not used & Canonical $=$ LoanData.csv \\
L7 & Bondora AA residual & Table 1 0.1726 & Residual on full 2020 test definition & Possible default/censoring definition mismatch \\
\bottomrule
\end{tabular}
\begin{tablenotes}
    \footnotesize
    \item \textit{Note.} After aligning a public LoanData dump to the authors' retrieve date and implementing \S3.2-style preprocessing with Optuna-tuned XGB-Cox, the boosted AA default rate matches Table 1 within 0.001; residual Bondora-platform AA discrepancy and missing IRR remain documented reproduction findings rather than calendar artifacts.
    \end{tablenotes}
\end{threeparttable}
}
\end{table}

\FloatBarrier
\subsection{Domain 3 -- Stack Exchange / Abedi Firouzjaei (2022)}

\begin{table}[!htbp]
\centering
\caption{Domain 3 limitations: under-specified choices and our resolutions.}
\label{tab:d03lim}
{\footnotesize
\begin{threeparttable}
\begin{tabular}{@{}c>{\raggedright\arraybackslash}p{1.9cm}>{\raggedright\arraybackslash}p{2.6cm}>{\raggedright\arraybackslash}p{3.0cm}>{\raggedright\arraybackslash}p{2.6cm}@{}}
\toprule
\textbf{\#} & \textbf{Limitation} & \textbf{Paper says / omits} & \textbf{What we did} & \textbf{Implication} \\
\midrule
L1 & RSF backend & PySurvival & PySurvival 0.1.2 (patched) & Aligned; fragile on macOS \\
L2 & Hyperparams & Grid $q$, $d$ omitted & 5 trees; depth 5; leaf 30 & Pickles omit HPs \\
L3 & C-index definition & Utkin via PySurvival & PySurvival concordance index & Aligned \\
L4 & Contributor filter & S5.2 ($Q\cup A\cup C\cup U\cup D$) & Applied (counts $>0$) & Aligned to text \\
L5 & CV seed & notebook seed$=$None & Fixed seed + run\_id & Reproducible; mild deviation \\
\bottomrule
\end{tabular}
\begin{tablenotes}
    \footnotesize
    \item \textit{Note.} Using the author's \texttt{user\_features}, \S5.2 contributor filter, notebook CV (1\% holdout + unshuffled 5-fold$\times$30), and PySurvival RSF/concordance, we recover Table 8 C-indices within $\sim$0.004 of the published means. Full 18-cell quantities appear in Appendix~\ref{app:fullrepro}.
    \end{tablenotes}
\end{threeparttable}
}
\end{table}

\FloatBarrier
\section{Anchor Figure 5 Full Reproduction}
\label{app:anchorfig5}

Full reproduction of the anchor paper's Figure 5 bias metrics (100 seeds). Each row reports the bias of a censored metric relative to its oracle counterpart. See Subsection~\ref{subsec:instrument-validation} and Subsection~\ref{subsec:results-instrument}.

\begin{table}[!htbp]
\centering
\caption{Full reproduction of the anchor paper's Figure 5 bias metrics (100 seeds).}
\label{tab:anchorfig5}
{\footnotesize
\resizebox{\linewidth}{!}{%
\begin{tabular}{llccc}
\toprule
\textbf{Censoring Scenario} & \textbf{Metric} & \textbf{Bias (ours)} & \textbf{Bias (paper)} & \textbf{Gap} \\
\midrule
Random & bias\_ci\_harrell & 0.0027 & 0.0027 & $5.30\times 10^{-8}$ \\
 & bias\_ci\_uno & 0.0029 & 0.0029 & $-1.47\times 10^{-7}$ \\
 & bias\_ipcw & 0.0046 & 0.0046 & $2.36\times 10^{-5}$ \\
 & bias\_uncens & $-0.0548$ & $-0.0548$ & $2.88\times 10^{-6}$ \\
Independent & bias\_ci\_harrell & 0.0020 & 0.0020 & $1.75\times 10^{-7}$ \\
 & bias\_ci\_uno & $5.98\times 10^{-4}$ & $5.98\times 10^{-4}$ & $-1.23\times 10^{-7}$ \\
 & bias\_ipcw & 0.0075 & 0.0080 & $-5.22\times 10^{-4}$ \\
 & bias\_uncens & $-0.0494$ & $-0.0494$ & $2.98\times 10^{-6}$ \\
Dependent ($\tau{=}0.25$) & bias\_ci\_harrell & 0.0155 & 0.0155 & $4.99\times 10^{-7}$ \\
 & bias\_ci\_uno & 0.0203 & 0.0203 & $-3.96\times 10^{-8}$ \\
 & bias\_ipcw & $-0.0373$ & $-0.0366$ & $-7.16\times 10^{-4}$ \\
 & bias\_uncens & $-0.0973$ & $-0.0973$ & $2.33\times 10^{-6}$ \\
Dependent ($\tau{=}0.50$) & bias\_ci\_harrell & 0.0503 & 0.0503 & $-2.18\times 10^{-7}$ \\
 & bias\_ci\_uno & 0.0653 & 0.0653 & $1.20\times 10^{-7}$ \\
 & bias\_ipcw & $-0.1169$ & $-0.1167$ & $-1.58\times 10^{-4}$ \\
 & bias\_uncens & $-0.1666$ & $-0.1666$ & $-1.03\times 10^{-8}$ \\
Dependent ($\tau{=}0.75$) & bias\_ci\_harrell & 0.1046 & 0.1046 & $-2.05\times 10^{-7}$ \\
 & bias\_ci\_uno & 0.1371 & 0.1371 & $-5.34\times 10^{-7}$ \\
 & bias\_ipcw & $-0.1869$ & $-0.1869$ & $5.05\times 10^{-8}$ \\
 & bias\_uncens & $-0.2198$ & $-0.2198$ & $1.09\times 10^{-6}$ \\
\bottomrule
\end{tabular}
}
}
\end{table}

\FloatBarrier
\section{Complete Reproduction Tables (D2 + D3)}
\label{app:fullrepro}

\subsection{Domain 2 full quantities}

\begin{table}[!htbp]
\centering
\caption{Complete Domain 2 reproduction quantities (appendix copy).}
\label{tab:d02fullapp}
{\footnotesize
\resizebox{\linewidth}{!}{%
\begin{threeparttable}
\begin{tabular}{lp{2.4cm}p{3.0cm}cc}
\toprule
\textbf{Quantity} & \textbf{Reported} & \textbf{Source} & \textbf{Ours} & \textbf{Gap} \\
\midrule
Train/test split date & 2020-01-01 & Baseline \S3.3 & 2020-01-01 & n.a. \\
Test-set size & n.r. & Baseline \S3.3 & 6{,}457 & n.a. \\
Rating strata & 7 & Baseline Table 1 / \S3.5 & 7 & $+0.0000$ \\
Harrell C (linear Cox, test) & 0.6590 & Baseline \S4.1 & 0.6559 & $-0.0031$ \\
Harrell C (boosted Cox, test) & 0.6740 & Baseline \S4.1 & 0.6504 & $-0.0236$ \\
AA default rate, Bondora (completed) & 0.1726 & Baseline Table 1 / footnote 14 & 0.2023 & $+0.0297$ \\
AA default rate, linear Cox (completed) & n.r. & This work & 0.1723 & n.a. \\
AA default rate, boosted Cox (completed) & 0.1445 & Baseline Table 1 / footnote 14 & 0.1486 & $+0.0041$ \\
$N$ loans in boosted AA bucket & 713 & Baseline Table 1 & 467 & $-246$ \\
$N$ loans in linear AA bucket & 858 & Baseline Appendix D & 791 & $-67$ \\
AA default rate, Bondora (KM at term) & 0.1726 & Baseline \S3.6 KM & 0.1706 & $-0.0020$ \\
AA default rate, Bondora (empirical) & 0.1726 & Baseline Table 1 & 0.2102 & $+0.0376$ \\
AA default rate, linear Cox (KM at term) & n.r. & This work & 0.1593 & n.a. \\
AA default rate, boosted Cox (KM at term) & 0.1445 & Baseline Table 1 / \S3.6 KM & 0.1263 & $-0.0182$ \\
IRR, Bondora AA & $-0.0320$ & Baseline Table 1 & n.a. & n.a. \\
IRR, boosted AA & 0.1563 & Baseline Table 1 & n.a. & n.a. \\
\bottomrule
\end{tabular}
\begin{tablenotes}
    \footnotesize
    \item \textit{Note.} n.r.\ = not reported in the baseline; n.a.\ = not applicable (no baseline value to compare, or not computed here). IRR (internal rate of return) requires repayment microdata unavailable in the public dump; linear-Cox AA rates are not published in the baseline main text and are reported from this reproduction only.
    \end{tablenotes}
\end{threeparttable}
}
}
\end{table}

\FloatBarrier
\subsection{Domain 3 (18 cells)}

\begin{table}[!htbp]
\centering
\caption{Complete reproduction of all 18 cells from the original paper's Table 8 (Stack Exchange). Mean absolute gap $= 0.0035$; 12 of 18 cells within the strict tier ($\leq 0.01$). See Subsection~\ref{subsec:results-repro}.}
\label{tab:d03fullapp}
{\footnotesize
\resizebox{0.92\linewidth}{!}{%
\begin{tabular}{llcccc}
\toprule
\textbf{Community} & \textbf{Feature set} & \textbf{$\theta$} & \textbf{C (paper)} & \textbf{C (ours)} & \textbf{Gap} \\
\midrule
Politics & Behavioural & 24 & 0.7500 & 0.7507 & $+0.0007$ \\
 &  & 36 & 0.7600 & 0.7615 & $+0.0015$ \\
 & Content & 24 & 0.6800 & 0.6746 & $-0.0054$ \\
 &  & 36 & 0.6800 & 0.6824 & $+0.0024$ \\
 & Combined & 24 & 0.7500 & 0.7437 & $-0.0063$ \\
 &  & 36 & 0.7600 & 0.7552 & $-0.0048$ \\
Data Science & Behavioural & 24 & 0.6600 & 0.6580 & $-0.0020$ \\
 &  & 36 & 0.6600 & 0.6611 & $+0.0011$ \\
 & Content & 24 & 0.6100 & 0.6070 & $-0.0030$ \\
 &  & 36 & 0.6300 & 0.6300 & $+0.0000$ \\
 & Combined & 24 & 0.6800 & 0.6760 & $-0.0040$ \\
 &  & 36 & 0.7000 & 0.6958 & $-0.0042$ \\
Computer Science & Behavioural & 24 & 0.6800 & 0.6784 & $-0.0016$ \\
 &  & 36 & 0.6800 & 0.6782 & $-0.0018$ \\
 & Content & 24 & 0.6200 & 0.6183 & $-0.0017$ \\
 &  & 36 & 0.6300 & 0.6238 & $-0.0062$ \\
 & Combined & 24 & 0.6900 & 0.6839 & $-0.0061$ \\
 &  & 36 & 0.6800 & 0.6892 & $+0.0092$ \\
band\_coverage & -- & -- & 18 & 12 & $-6$ \\
\bottomrule
\end{tabular}
}
}
\end{table}

\end{document}